\documentclass{article} %
\usepackage[preprint]{iclr2027_conference}
\usepackage[T1]{fontenc}
\usepackage{times}

\usepackage{amsmath,amsfonts,bm}

\def\eqref#1{equation~\ref{#1}}

\def\1{\bm{1}}

\DeclareMathAlphabet{\mathsfit}{\encodingdefault}{\sfdefault}{m}{sl}
\SetMathAlphabet{\mathsfit}{bold}{\encodingdefault}{\sfdefault}{bx}{n}

\usepackage[hidelinks]{hyperref}
\usepackage{url}
\usepackage{booktabs}
\usepackage{multirow}
\usepackage{graphicx}
\usepackage{colortbl}
\usepackage{tabularx}
\usepackage{wrapfig}
\usepackage{algorithm}
\usepackage{algorithmic}
\usepackage{subcaption}
\usepackage{enumitem}
\usepackage{pifont}
\usepackage{float}

\usepackage[most]{tcolorbox}
\usepackage{xcolor}

\definecolor{lightblue}{rgb}{0.22,0.45,0.70}

\tcbset{
  aibox/.style={
    width=\linewidth,
    top=10pt,
    bottom=6pt,
    left=8pt,
    right=8pt,
    colback=blue!4!white,
    colframe=black,
    colbacktitle=black,
    coltitle=white,
    fonttitle=\bfseries,
    enhanced,
    center,
    attach boxed title to top left={yshift=-0.10in,xshift=0.15in},
    boxed title style={boxrule=0pt,colframe=white},
  }
}

\newtcolorbox{AIbox}[2][]{aibox,title=#2,#1}

\definecolor{darkred}{rgb}{0.5, 0.0, 0.0}
\definecolor{obsborder}{RGB}{41, 128, 185}   %
\definecolor{obsbg}{RGB}{245, 249, 253}       %
\definecolor{takeawayborder}{RGB}{39, 174, 96} %
\definecolor{takeawaybg}{RGB}{242, 249, 244}   %

\newcolumntype{C}{>{\centering\arraybackslash}X}

\title{Policy Plasticity Matters in \\Offline-to-Online Reinforcement Learning: \\Refitting Offline Policies \\for Online Adaptation}

\author{
  Yuheng Huang\thanks{These authors contributed equally to this work.}\\
    Zhejiang University
  \And
  Yunpeng Qing\footnotemark[1]\kern0.5em\thanks{Project lead.}\\
    Zhejiang University
  \And
  Yixiao Chi\\
    Carnegie Mellon University
  \And
  Yilun Kong\\
    Nanyang Technological University
  \And
  Changqing Zou\thanks{Corresponding author.}\\
    Zhejiang University \& Zhejiang Lab
}

\begin{document}

\maketitle

\begin{abstract}
Offline-to-Online Reinforcement Learning~(O2O RL) has emerged as a practical paradigm that pre-trains the policy using static offline datasets and subsequently adapts the policy through online interactions. 
Existing O2O methods primarily address the transition through value calibration, while generally treating the offline-trained policy as a given initialization.
We instead study O2O adaptation from the perspective of network plasticity, asking whether the offline-trained policy remains sufficiently adaptable for online learning.
Controlled experiments show that prolonged optimization on static offline data progressively reduces network plasticity even after offline performance has largely saturated, and that lower plasticity is associated with weaker subsequent online improvement.
Motivated by these observations, we propose REstoring plasticity via Fresh Initialization and policy Transfer~(REFIT), a lightweight model-level method for the O2O transition. 
Before online fine-tuning, REFIT distills the offline policy into a freshly initialized student while temporarily freezing a random subset of student units, transferring the learned offline behavior to a more plastic policy initialization.
Extensive experiments on D4RL and OGBench demonstrate that REFIT consistently achieves higher aggregate performance than existing O2O plug-in methods across both Cal-QL and IQL backbones, while plasticity diagnostics and ablations provide further evidence of restored network plasticity.

\end{abstract}

\section{Introduction}

Deep Reinforcement Learning~(DRL) has achieved strong performance across diverse real-world applications, such as robot control~\citep{qing2024a2po, qing2025bitrajdiff} and power grid control~\citep{xu2024temporal, chen2025powerformer}.
To learn such effective policies, DRL typically relies on either collecting experience through online interaction or leveraging previously collected datasets.
However, online interaction can be costly and unsafe~\citep{lillicrap_continuous_2016,haarnoja_soft_2018}, while learning solely from fixed datasets is limited by insufficient data distribution coverage and the challenge of out-of-distribution~(OOD) actions~\citep{fujimoto2019off,kostrikov2021offline,kumar2020conservative}.
Offline-to-Online~(O2O) RL bridges these two paradigms by pre-training on offline data and subsequently improving the policy through online interaction, thereby reducing the reliance on costly exploration and enabling adaptation beyond the coverage of the offline data~\citep{zhou2025efficient,nakamoto2023cal,zhang2023policy}.
 
\begin{figure}[t!]
    \centering
    \begin{subfigure}[t]{0.32\textwidth}
        \centering
        \includegraphics[width=\textwidth]{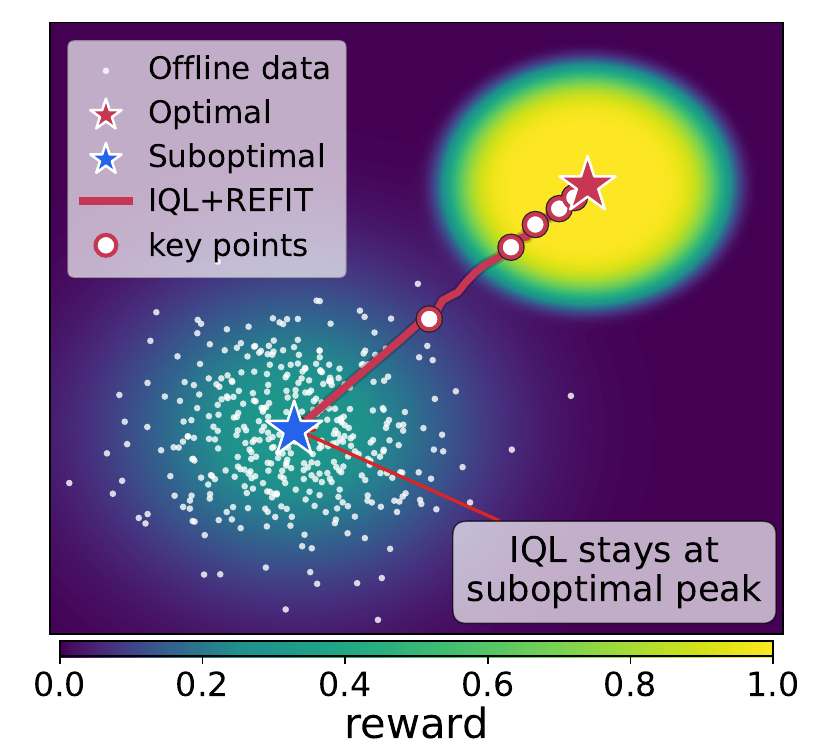}
        \caption{Environment setup}
        \label{fig:intro_toy_a}
    \end{subfigure}
    \hfill
    \begin{subfigure}[t]{0.32\textwidth}
        \centering
        \includegraphics[width=\textwidth]{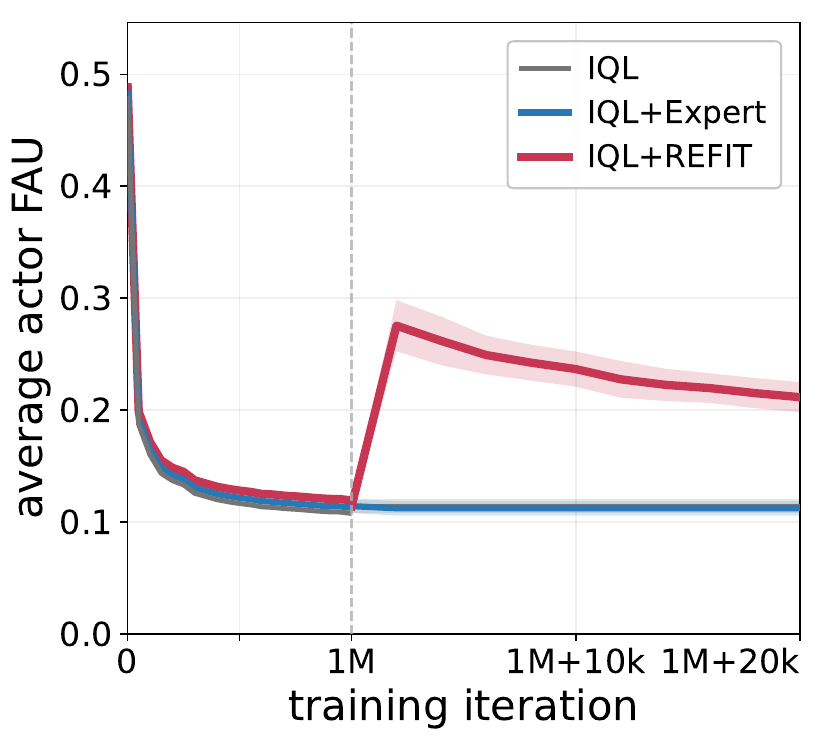}
        \caption{Actor FAU}
        \label{fig:intro_toy_b}
    \end{subfigure}
    \hfill
    \begin{subfigure}[t]{0.32\textwidth}
        \centering
        \includegraphics[width=\textwidth]{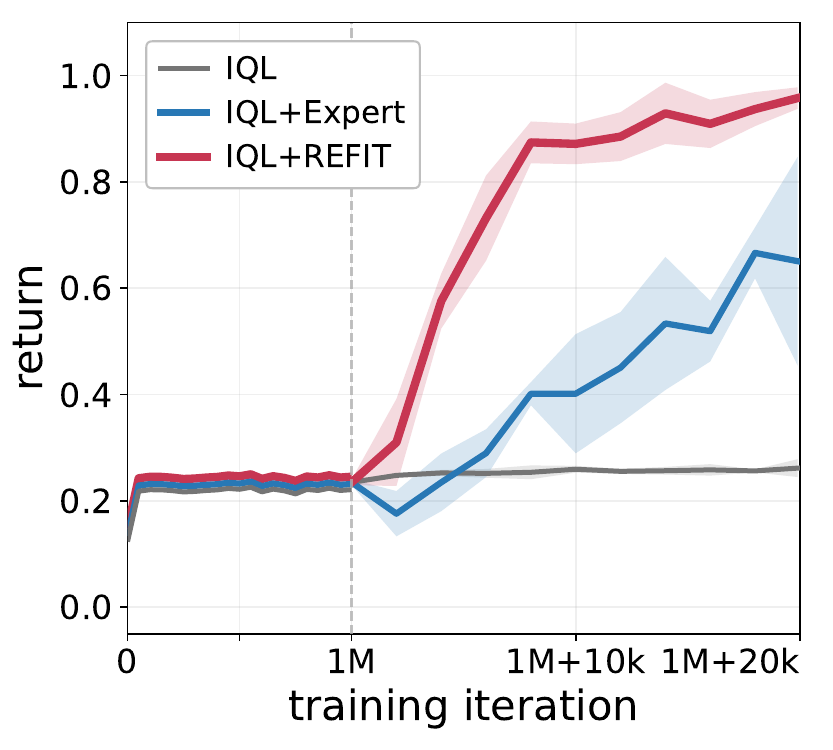}
        \caption{Learning curve}
        \label{fig:intro_toy_c}
    \end{subfigure}
    \caption{
        Network plasticity and online adaptation in a controlled O2O setting. The dashed line marks the transition from offline pre-training to online fine-tuning.
        (a)~The bimodal reward landscape, offline data distribution, and policy trajectories.
        (b)~Actor FAU throughout offline training and subsequent online fine-tuning.
        (c)~Learning curve throughout training.
    }
    \label{fig:intro_toy_example}
\end{figure}

Despite these promising results, O2O RL often suffers from limited asymptotic performance during online fine-tuning~\citep{zhang2024perspective, luo2023finetuningofflinereinforcementlearning}. 
To tackle this issue, existing methods generally take the offline-trained policy as a given initialization and focus on recalibrating value estimates under the offline-to-online shift, alleviating the excessive conservatism inherited from offline training~\citep{nakamoto2023cal, luo2024optimistic, shin2025online}.
These methods implicitly assume that the offline-trained policy remains sufficiently plastic to benefit from subsequent online updates. 
However, we argue that the network plasticity of the offline-trained policy can also be a critical bottleneck to subsequent online improvement. 
Specifically, the loss of plasticity during offline training can cause the network to become rigid, severely hindering its ability to move beyond the suboptimal behavior learned offline and adapt toward new behavior patterns that yield higher returns.
To exemplify the above issue, we design a controlled 2D continuous bandit experiment and track the Fraction of Active Units~(FAU) as a proxy for network plasticity.
As shown in Figure~\ref{fig:intro_toy_a}, the offline data cover only the suboptimal peak. 
During online fine-tuning, IQL remains near the suboptimal peak, while even with the guidance of expert data, the IQL+Expert variant can only improve slowly.
Their low actor FAU~(Figure~\ref{fig:intro_toy_b}) suggests that insufficient network plasticity, rather than data coverage alone, can hinder the offline-trained policy from adapting toward better behaviors.

\vspace{0ex}

Motivated by these observations, we propose \textbf{REstoring plasticity via Fresh Initialization and policy Transfer~(REFIT)} for O2O RL.
Unlike prior methods that focus on value correction~\citep{nakamoto2023cal,luo2024optimistic,shin2025online} to improve online learning efficiency, REFIT restores the plasticity of the offline-trained policy before online fine-tuning, allowing it to better adapt to the online distribution while preserving the offline prior.
Specifically, REFIT transfers the offline behavioral prior by distilling the teacher policy into a freshly initialized student network. 
Crucially, by temporarily freezing a random subset of units in the student network during this process, REFIT preserves a subset of unoptimized parameters that preserve high plasticity, facilitating rapid adaptation during the subsequent online phase.
As shown in Figure~\ref{fig:intro_toy_b} and Figure~\ref{fig:intro_toy_c}, even when the offline policy converges to a highly suboptimal solution, REFIT uses distillation to retain the offline prior while restoring network plasticity, moving beyond its offline behavior and continuing to improve during online fine-tuning. 

\textbf{Our contributions} are summarized as follows:
\begin{itemize}
    \item We identify an overlooked issue in O2O RL: 
    Offline optimization can reduce the plasticity of the policy network, hindering its ability to continue adapting during online fine-tuning.

    \item We propose REFIT, a simple plug-in method that distills the offline policy into a freshly initialized student policy while temporarily freezing a random subset of student units. 
    This procedure transfers the learned offline behavioral prior to a more plastic network, enabling continued adaptation during online fine-tuning.
    
    \item We demonstrate the effectiveness of REFIT through extensive experiments on D4RL and OGBench across two O2O RL backbones, showing that it consistently outperforms existing plug-in O2O RL methods and achieves stronger performance across various settings.
\end{itemize}

\section{Related Works}
\label{sec:related_works}

\subsection{Offline-to-Online Reinforcement Learning}
O2O RL first pre-trains a policy on a static dataset, and then fine-tunes it through online environment interactions~\citep{nair2020awac, nakamoto2023cal, fujimoto2019off, fujimoto2021minimalist}. 
It combines the strong initialization provided by offline RL with the continual improvement enabled by online interactions.
Early O2O methods directly extend conservative offline RL algorithms to the online phase, including Implicit Q-Learning~\citep{kostrikov2021offline} and Conservative Q-Learning~\citep{kumar2020conservative}. 
More recent methods have dived deep into the inner mechanisms of O2O RL such as recalibrating the value function to mitigate value underestimation~\citep{nakamoto2023cal} and reducing distribution mismatch via a warm-up phase~\citep{zhou2025efficient}. 
However, existing O2O methods primarily focus on these algorithmic challenges, while offline-induced plasticity degradation has received comparatively limited attention.
In contrast, our method explicitly restores network plasticity during O2O adaptation, leading to substantially improved performance.

\subsection{Plasticity in Reinforcement Learning}

Plasticity refers to the ability of a neural network to continually adapt to new data distributions~\citep{klein2024plasticity, dohare2021continual}. 
Recent studies have shown that deep RL agents progressively suffer from degrading plasticity, limiting their ability to adapt to new distributions despite having sufficient model capacity~\citep{abbas2023loss, lyle2022understanding, wang2026dual}. 
Existing work attributes this phenomenon to the non-stationarity of RL training and measures it through symptoms such as dormant neurons, representation collapse, and abnormal parameter or gradient dynamics~\citep{sokar2023dormant, klein2024plasticity, zhou2025stay}. 
To mitigate plasticity loss, prior methods have explored weight resets~\citep{nikishin2022primacy}, normalization~\citep{lyle2023understanding} and regularization techniques~\citep{kumar2024maintainingplasticitycontinuallearning}, as well as architectural designs that explicitly inject plasticity into the critic during training~\citep{nikishin2023deep}. 
Meanwhile, recent work~\citep{kong2024efficient} has realized that primacy bias can also be a key factor limiting subsequent O2O adaptation. 
However, direct assessment of policy plasticity during the O2O pipeline and corresponding solutions specifically targeting its restoration remain underexplored.
To fill this gap, we first systematically characterize offline-induced plasticity loss and propose REFIT to restore network plasticity at the O2O transition by distilling the offline teacher policy to an initialized and partially frozen online student policy.

\section{Preliminaries}
\label{sec:preliminaries}

\paragraph{Offline-to-Online Reinforcement Learning}
Traditional RL can be viewed as a Markov Decision Process (MDP)~\citep{puterman1990markov} paradigm, defined as a tuple $\mathcal{M} = (\mathcal{S}, \mathcal{A}, P, r, d_0, \gamma)$, where $\mathcal{S}$ is the state space, $\mathcal{A}$ is the action space, $P(s' \mid s, a)$ represents the transition dynamics, $r(s, a)$ is the reward function, $d_0(s)$ is the initial state distribution, and $\gamma \in [0, 1)$ denotes the discount factor. 
The objective of RL is to learn a policy $\pi(a \mid s)$ that maximizes the expected discounted return,
$\pi^* = \arg\max_{\pi}\mathbb{E}_{s_0\sim d_0,\;a_t\sim\pi(\cdot\mid s_t),\;s_{t+1}\sim P(\cdot\mid s_t,a_t)}
\left[\sum_{t=0}^{\infty}\gamma^t r(s_t,a_t)\right]$.
The policy is evaluated by the state-value function $V^\pi(s) = \mathbb{E}_{\pi} [ \sum_{t=0}^{\infty} \gamma^t r(s_t, a_t) \mid s_0 = s ]$ and the action-value function $Q^\pi(s, a) = \mathbb{E}_{\pi} [ \sum_{t=0}^{\infty} \gamma^t r(s_t, a_t) \mid s_0 = s, a_0 = a ]$.
As for O2O RL, the agent is given a dataset $\mathcal{D}_{\text{off}} = \{(s_t, a_t, r_t, s_{t+1})\}_{t=1}^N$ collected by a behavior policy $\pi_\beta$. 
The agent leverages $\mathcal{D}_{\text{off}}$ to pre-train the value functions $Q_{\text{offline}}$, $V_{\text{offline}}$ and policy $\pi_{\text{offline}}$, which then serve as the starting point for online interaction and optimization.

\paragraph{Model Plasticity}
Network plasticity~\citep{lyle2023understanding, dohare2021continual, nikishin2022primacy} refers to the capacity of a neural network to continually adapt to new data distributions, which can be quantitatively monitored through activation patterns, parameter scales, and gradient behaviors.
In this work, we primarily focus on the \textit{Fraction of Active Units (FAU)}~\citep{ma2023revisiting} as a key metric for diagnosing network plasticity. Additional plasticity metrics are provided in Appendix~\ref{appendix:plasticity_metrics}.
Specifically, for a hidden layer $l$ with width $H_l$, we define the layer-wise FAU as well as the network FAU under a state distribution $\mathcal{B}$ as
\begin{equation}
    \Phi_l =
    \mathbb{E}_{s \sim \mathcal{B}}
    \left[
    \frac{1}{H_l}\sum_{n=1}^{H_l}
    \mathbb{I}\!\left(h_n^{(l)}(s)>0\right)
    \right],
    \qquad
    \Phi_{\mathrm{network}} = \frac{1}{L}\sum_{l=1}^{L}\Phi_l,
\end{equation}
where $h_n^{(l)}(s)$ denotes the pre-activation of unit $n$ in layer $l$, $L$ is the number of hidden layers, and $\mathbb{I}(\cdot)$ is the indicator function. 
This statistic measures the fraction of hidden units with nonzero post-ReLU activation.

\section{Understanding and Mitigating Plasticity Loss in O2O RL}
\label{sec:methodology}
\vspace{0ex}

To understand what limits online adaptation in O2O RL, we examine the plasticity of the offline-trained policy.
Specifically, we first provide controlled empirical evidence that offline over-optimization progressively reduces network plasticity and suppresses subsequent online adaptation.
This observation motivates \textbf{Plasticity Restoration} as a key design principle for O2O RL.
Following this principle, we propose REstoring plasticity via Fresh Initialization and policy Transfer, termed \textbf{REFIT}, a plug-in method orthogonal to existing O2O RL algorithms that transfers the behavioral prior of the offline teacher policy to a freshly initialized online student policy through randomized partial freezing and behavioral distillation.

\vspace{0ex}
\subsection{Empirical Analysis: Plasticity Loss in O2O RL}
\label{plasticity_toy}

\begin{figure}[htbp]
    \centering
    \includegraphics[width=0.60\textwidth]{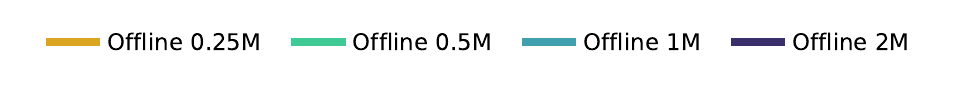}
    \vspace{0ex}

    \begin{subfigure}[b]{0.48\textwidth}
        \centering
        \includegraphics[width=\textwidth]{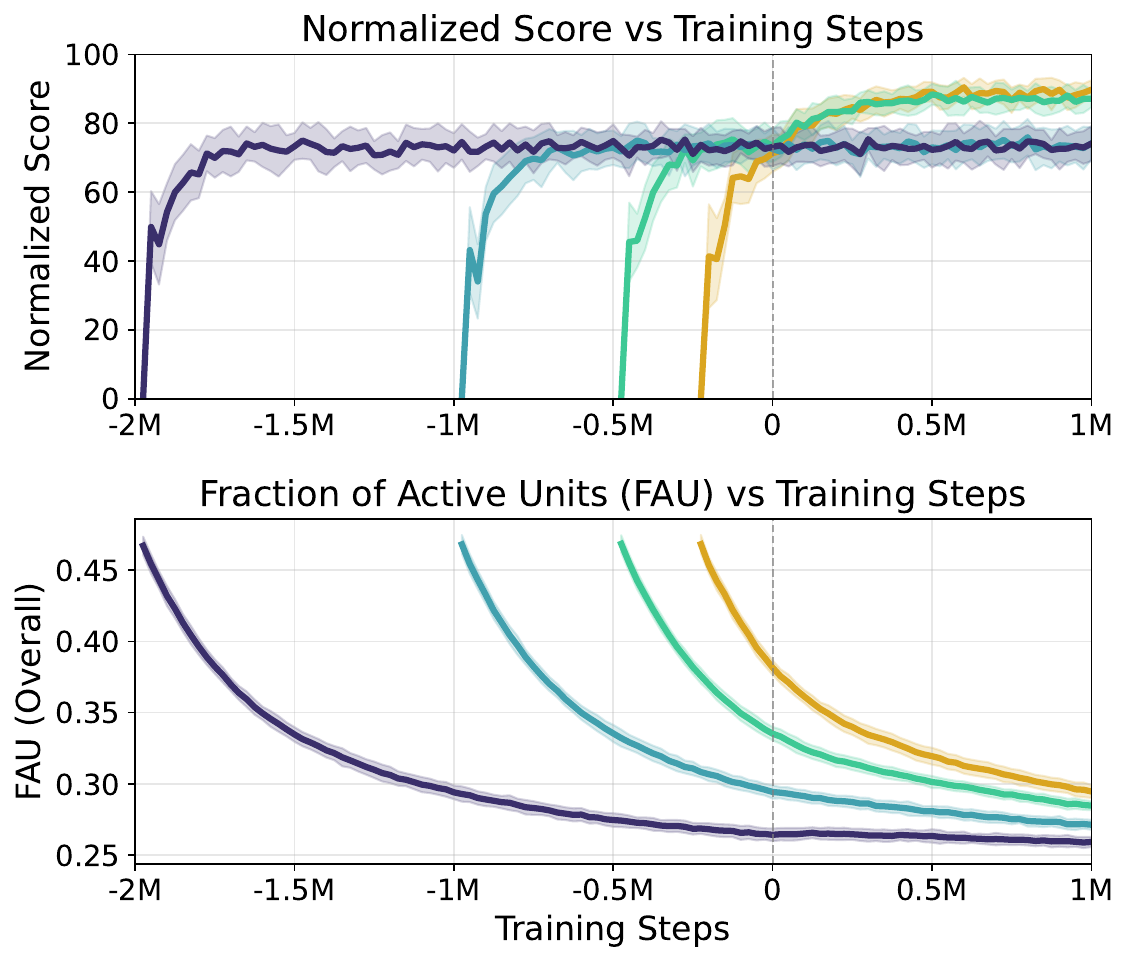}
        \caption{Performance and FAU of O2O IQL}
        \label{fig:plasticity_left}
    \end{subfigure}
    \hfill
    \begin{subfigure}[b]{0.48\textwidth}
        \centering
        \includegraphics[width=\textwidth]{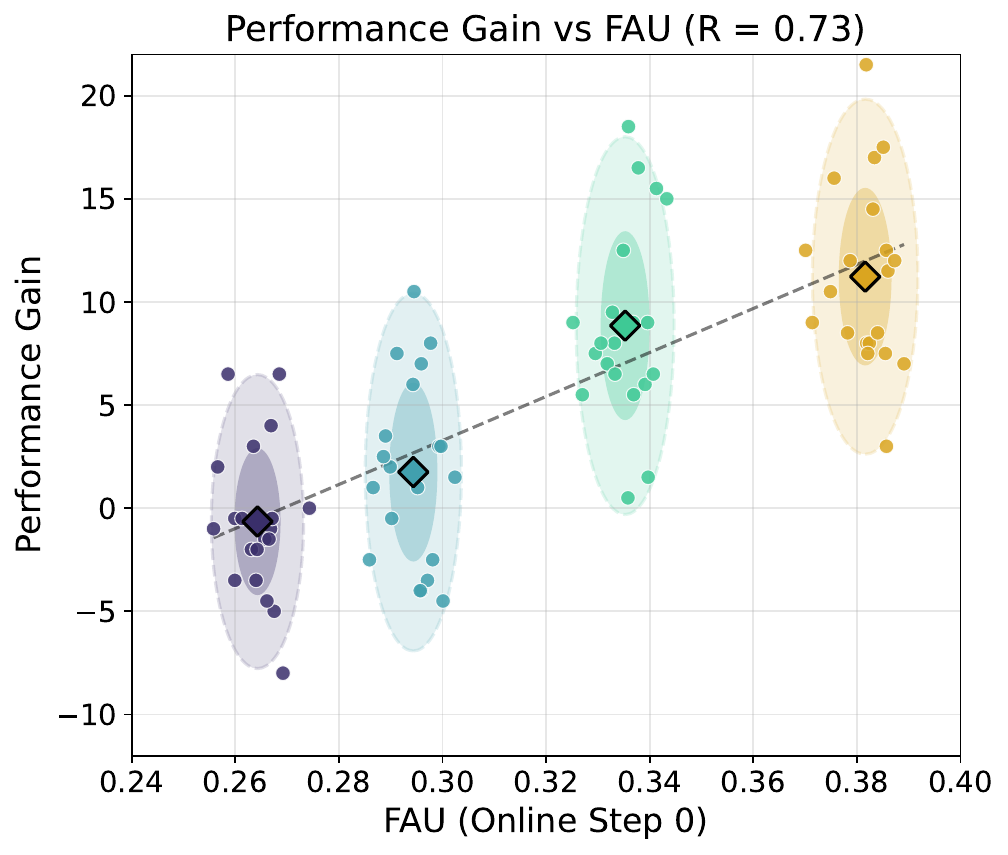}
        \caption{Correlation between FAU and performance gain}
        \label{fig:plasticity_right}
    \end{subfigure}

    \caption{
    Offline over-optimization significantly degrades model plasticity, and the plasticity metric FAU is shown to be positively correlated with online performance improvements.
    (a) Overall performance and FAU curves across varying offline pre-training budgets.
    (b) Correlation between online initial policy FAU and performance improvement.
    }
    \label{fig:plasticity_all}
\end{figure}

To investigate the relationship between offline training, network plasticity, and subsequent online adaptation, we conduct a controlled empirical study based on IQL~\citep{kostrikov2021offline} on the D4RL \texttt{antmaze-medium-play} environment, as shown in Figure~\ref{fig:plasticity_all}.
To disentangle the effect of network plasticity from that of offline policy quality at the start of online fine-tuning, we pre-train identical policies with four offline training budgets: $0.25\text{M}$, $0.5\text{M}$, $1\text{M}$, and $2\text{M}$ gradient steps.
These checkpoints attain comparable near-saturated offline performance while exhibiting distinct levels of network plasticity, thereby allowing us to study how plasticity affects subsequent online adaptation. 
During the whole process, we track the normalized score to measure adaptation performance and FAU~\citep{ma2023revisiting} as a proxy for network plasticity. 
A decline in FAU reflects a reduced fraction of active units, further indicating increased plasticity loss and diminished adaptability.
All the above experiments are conducted over $20$ random seeds.
Given the results shown in Figure~\ref{fig:plasticity_all}, we have the following observations. 

\vspace{0ex}

\vspace{0ex}

\begin{AIbox}{Observation 1: Offline Over-Optimization Leads to Plasticity Degradation.}
Continued optimization on static offline data progressively reduces FAU even after offline performance has largely saturated, indicating a gradual plasticity loss before online fine-tuning.
\end{AIbox}

As shown in Figure~\ref{fig:plasticity_left}, all of the settings achieve comparable offline performance, while FAU consistently decreases with continued offline optimization. 
This indicates that network plasticity can degrade even after offline performance has largely saturated. 
The results suggest that over-optimization on a static offline dataset may progressively constrain the diversity of network activations, reducing the fraction of active units and potentially limiting the capacity of the model to adapt to novel distributions encountered during online fine-tuning.

\begin{AIbox}{Observation 2: Higher Plasticity Correlates with Stronger Online Adaptation.}
Policies with higher FAU exhibit larger online performance gains, whereas lower-FAU policies adapt more slowly and tend to plateau earlier.
\end{AIbox}

During online fine-tuning, the policy must adapt to online interaction data beyond the fixed offline distribution.
As shown in Figure~\ref{fig:plasticity_left}, online improvement steadily diminishes as the offline training budget increases from $0.25\text{M}$ to $2\text{M}$ steps. In particular, the $1\text{M}$- and $2\text{M}$-step policies show little further improvement during online fine-tuning, consistent with weaker adaptability at lower plasticity.
This relationship is further reflected in Figure~\ref{fig:plasticity_right}, where mean FAU is positively correlated with online performance gain, linking higher retained plasticity to stronger subsequent online adaptation.

\paragraph{Design Implication: Plasticity Restoration for O2O RL}

These empirical observations suggest that offline optimization can progressively reduce policy plasticity and weaken subsequent online adaptation.
This motivates treating plasticity restoration as an explicit design objective at the O2O transition.
Guided by this insight, we design REFIT to transfer the behavioral prior from the offline teacher policy to a freshly initialized student policy, restoring the capacity of the agent for subsequent online adaptation without discarding the learned offline behavior.

\begin{wrapfigure}{r}{0.52\textwidth}
    \centering
    \vspace{-16pt}
    \includegraphics[width=\linewidth]{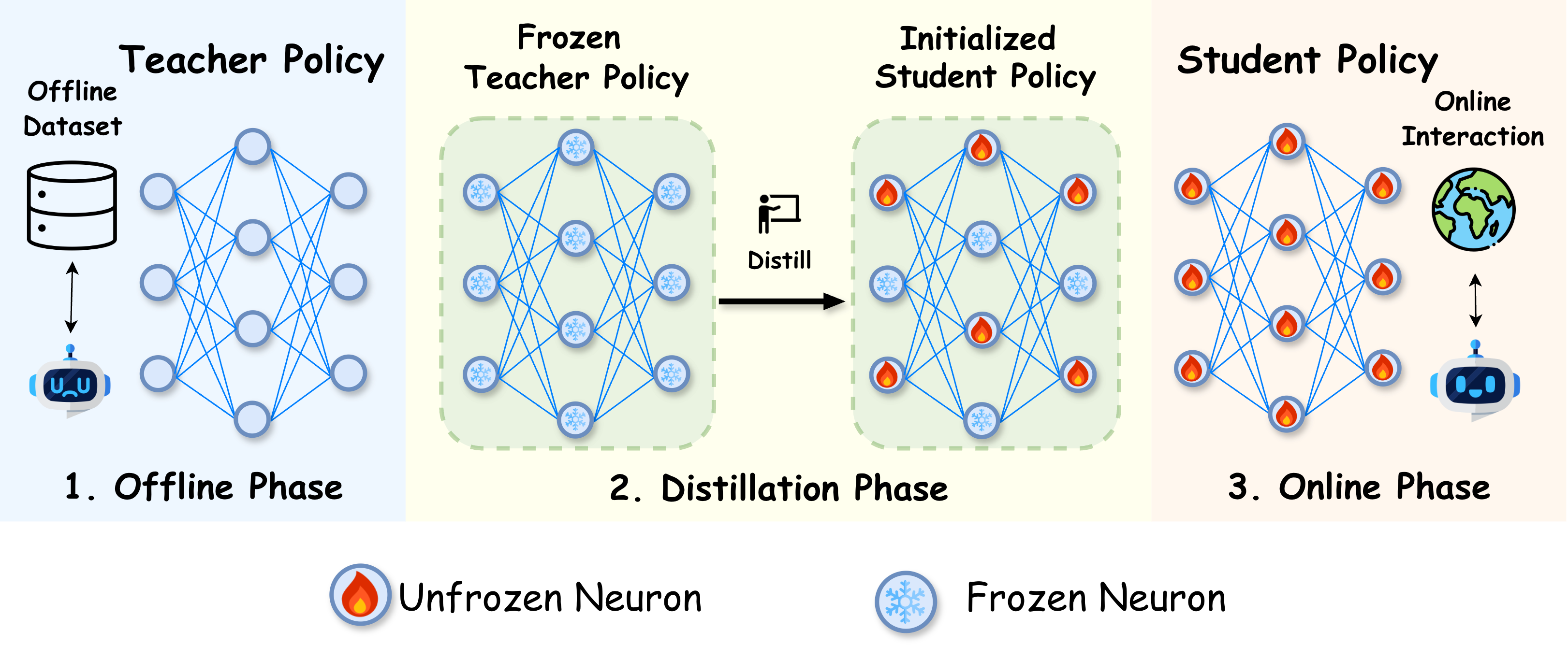}
    \vspace{-18pt}
    \caption{Schematic overview of REFIT.}
    \label{fig:REFIT_illustrate}
    \vspace{-12pt}
\end{wrapfigure}

\subsection{REFIT for Plasticity Restoration}
\label{sec:REFIT_method}

Motivated by the above analysis, we introduce \textbf{REFIT}, a lightweight model-level method for restoring policy plasticity at the O2O transition. 
Rather than directly continuing online optimization from the offline-trained policy parameters, REFIT instantiates a freshly initialized student policy and transfers the behavior learned by the offline teacher through policy distillation. 
During distillation, a random subset of student units is temporarily frozen at their initial values, allowing the student to acquire the offline behavior while retaining part of its freshly initialized structure. 
Once distillation is complete, the frozen units are released, and the resulting online student policy undergoes online fine-tuning with all parameters trainable.

\paragraph{Randomized Partial Freezing.}
Before online fine-tuning, REFIT instantiates a fresh online student policy $\pi_{\text{student}}$ from the standard initialization distribution and randomly partitions its parameters \(\theta_{\text{student}}\) into two disjoint subsets:
\begin{equation}
    \theta_{\text{student}} = \theta_{\text{train}} \cup \theta_{\text{freeze}},
    \quad
    \theta_{\text{train}} \cap \theta_{\text{freeze}} = \emptyset.
\end{equation}

Specifically, REFIT randomly selects a fraction $\rho$ of units in each dense layer and freezes their associated parameters at their initial values throughout distillation, forming $\theta_{\text{freeze}}$.
The remaining parameters $\theta_{\text{train}}$ are optimized to reproduce the behavior of the offline teacher policy.
By leaving part of the freshly initialized student unoptimized during behavioral transfer, partial freezing allows the trainable parameters to acquire the offline behavioral prior through distillation while preserving a freshly initialized parameter subset for subsequent online adaptation.

\paragraph{Plasticity-Preserving Distillation.}

Given the partition above, REFIT transfers the behavior of the offline teacher policy $\pi_{\text{teacher}}$ to the student using the offline dataset $\mathcal{D}_{\text{off}}$.
During distillation, only the trainable subset $\theta_{\text{train}}$ is optimized, while $\theta_{\text{freeze}}$ remains fixed at its initial values:
\begin{equation}
    \mathcal{L}_{\text{distill}}(\theta_{\text{train}})
    =
    \mathbb{E}_{s \sim \mathcal{D}_{\text{off}}}
    \left[
        \left\|
        \pi_{\theta_{\text{student}}}(s)
        -
        \pi_{\text{teacher}}(s)
        \right\|_2^2
    \right].
    \label{eq:distill_formula}
\end{equation}
This objective allows the trainable parameters to recover the behavior learned by the offline teacher, while the frozen subset remains unaffected by distillation and retains its fresh initialization.
REFIT is applied only to the policy network; the critic and value networks are directly inherited from the offline phase.
After distillation, the freeze mask is removed and all student parameters are set to trainable for subsequent online fine-tuning.
In this way, REFIT combines fresh initialization, behavioral distillation, and temporary partial freezing to transfer the offline policy behavior while providing a more plastic initialization for online adaptation.
Pseudocode of our algorithm is provided in Appendix~\ref{appx:REFIT-algorithm}.

\section{Experiments}
\label{exp}

Our experiments aim to answer the following questions: whether REFIT (1) improves the online performance of original O2O RL algorithms over other plug-in methods (Section~\ref{exp:main_exp}), 
(2) restores network plasticity during online fine-tuning (Section~\ref{exp:plasticity_exp}), 
(3) benefits from each design component (Section~\ref{exp:ablation_exp}), 
(4) remains effective without manually tuning the offline training budget (Section~\ref{exp:offline_budget}), 
and (5) introduces only modest computational overhead (Appendix~\ref{appendix:time}).

\begin{table}[htbp]
    \centering
    \caption{Performance of our REFIT and baselines on AntMaze, MuJoCo, and OGBench, averaged over five random seeds. \textbf{Bold} and \underline{underlined} values denote the best and second-best results within each backbone, respectively.}
    \label{tab:performance_comparison}
    \small
    \setlength{\tabcolsep}{4pt} 
    \resizebox{\textwidth}{!}{
    \begin{tabular}{l|ccc|ccc}
        \toprule
        \multirow{2}{*}{Environment} & \multicolumn{3}{c|}{IQL~\citep{kostrikov2021offline}} & \multicolumn{3}{c}{Cal-QL~\citep{nakamoto2023cal}} \\
        \cmidrule(lr){2-4} \cmidrule(lr){5-7}
        & Baseline & OPT & Ours (REFIT) & Baseline & PARS & Ours (REFIT) \\
        \midrule
        antmaze-m-p      & 76.2 $\pm$ 1.5 & \textbf{88.0} $\pm$ 3.0 & \underline{86.6} $\pm$ 4.8 & 92.2 $\pm$ 2.1 & \underline{92.4} $\pm$ 3.0 & \textbf{95.4} $\pm$ 1.7 \\
        antmaze-m-d      & 71.6 $\pm$ 2.7 & \underline{88.8} $\pm$ 1.2 & \textbf{89.2} $\pm$ 2.7 & \underline{93.2} $\pm$ 0.8 & 89.6 $\pm$ 2.9 & \textbf{96.2} $\pm$ 0.8 \\
        antmaze-l-p      & 46.0 $\pm$ 2.8 & \underline{61.6} $\pm$ 5.1 & \textbf{63.8} $\pm$ 8.3 & 70.6 $\pm$ 7.5 & \textbf{78.6} $\pm$ 3.6 & \underline{78.2} $\pm$ 5.0 \\
        antmaze-l-d      & 50.6 $\pm$ 5.7 & \underline{63.6} $\pm$ 3.6 & \textbf{65.2} $\pm$ 4.0 & 53.6 $\pm$ 17.4 & \underline{79.2} $\pm$ 12.4 & \textbf{80.2} $\pm$ 5.0 \\
        \midrule
        \textbf{AntMaze total} & 244.4 & \underline{302.0} & \textbf{304.8} & 309.6 & \underline{339.8} & \textbf{350.0} \\
        \midrule
        \midrule
        halfcheetah-r    & 13.8 $\pm$ 3.9  & \underline{33.2} $\pm$ 11.8 & \textbf{46.3} $\pm$ 1.4 & \underline{33.4} $\pm$ 3.8 & 27.0 $\pm$ 2.4 & \textbf{39.2} $\pm$ 3.5 \\
        hopper-r         & 8.0 $\pm$ 0.3   & \underline{11.4} $\pm$ 0.5  & \textbf{15.7} $\pm$ 4.1 & 8.3 $\pm$ 1.7  & \underline{15.0} $\pm$ 0.8 & \textbf{31.2} $\pm$ 0.9 \\
        walker2d-r       & 7.1 $\pm$ 1.0   & \underline{10.0} $\pm$ 1.0  & \textbf{10.7} $\pm$ 1.1 & 11.5 $\pm$ 1.5 & \underline{11.7} $\pm$ 0.4 & \textbf{13.5} $\pm$ 3.8 \\
        halfcheetah-m    & 48.1 $\pm$ 0.1  & \textbf{54.3} $\pm$ 2.7  & \underline{50.0} $\pm$ 0.3 & \textbf{65.3} $\pm$ 3.0 & \underline{64.2} $\pm$ 1.1 & 51.1 $\pm$ 0.9 \\
        hopper-m         & 63.9 $\pm$ 3.2  & \textbf{91.6} $\pm$ 12.1 & \underline{79.8} $\pm$ 3.9 & \underline{76.3} $\pm$ 6.1 & 71.7 $\pm$ 1.8 & \textbf{82.2} $\pm$ 2.2 \\
        walker2d-m       & 71.9 $\pm$ 2.5  & \underline{78.9} $\pm$ 3.0  & \textbf{87.9} $\pm$ 0.6 & 83.4 $\pm$ 2.2  & \underline{88.8} $\pm$ 0.7 & \textbf{97.3} $\pm$ 2.9 \\
        halfcheetah-m-r  & 43.9 $\pm$ 0.2  & \underline{46.9} $\pm$ 2.8  & \textbf{47.1} $\pm$ 0.4 & \textbf{58.9} $\pm$ 0.8  & 54.4 $\pm$ 0.3 & \underline{55.8} $\pm$ 0.7 \\
        hopper-m-r       & 77.5 $\pm$ 18.1 & \textbf{104.8} $\pm$ 1.4 & \underline{102.7} $\pm$ 0.4 & 68.9 $\pm$ 7.8  & \underline{85.7} $\pm$ 4.6 & \textbf{87.6} $\pm$ 7.5 \\
        walker2d-m-r     & 74.8 $\pm$ 2.8  & \underline{93.6} $\pm$ 1.1  & \textbf{96.9} $\pm$ 2.2 & 86.7 $\pm$ 8.7  & \textbf{101.7} $\pm$ 1.3 & \underline{97.9} $\pm$ 5.8 \\
        \midrule
        \textbf{MuJoCo total} & 409.0 & \underline{524.7} & \textbf{537.1} & 492.7 & \underline{520.2} & \textbf{555.8} \\
        \midrule
        \midrule
        cube-single-play & 70.7 $\pm$ 9.7 & \underline{73.0} $\pm$ 9.4 & \textbf{81.1} $\pm$ 5.3 & 15.8 $\pm$ 16.2 & \underline{63.2} $\pm$ 1.9 & \textbf{79.6} $\pm$ 2.2 \\
        puzzle-3x3-play  & 25.3 $\pm$ 4.6 & \underline{27.7} $\pm$ 6.4 & \textbf{98.6} $\pm$ 1.8 & 1.4 $\pm$ 3.1 & \textbf{17.0} $\pm$ 9.7 & \underline{9.0} $\pm$ 8.3 \\
        scene-play       & 76.7 $\pm$ 5.7 & \underline{77.0} $\pm$ 3.0 & \textbf{96.9} $\pm$ 2.4 & \textbf{11.8} $\pm$ 11.6 & 1.4 $\pm$ 1.1 & \underline{6.0} $\pm$ 5.4 \\
        \midrule
        \textbf{OGBench total} & 172.7 & \underline{177.7} & \textbf{276.6} & 29.0 & \underline{81.6} & \textbf{94.6} \\
        \midrule
        \midrule
        \rowcolor[gray]{0.9} \textbf{All total} & 826.1 & \underline{1004.4} (+21.6\%) & \textbf{1118.5} (+35.4\%) & 831.3 & \underline{941.6} (+13.3\%) & \textbf{1000.4} (+20.3\%) \\
        \bottomrule
    \end{tabular}%
    } %
\end{table}

\paragraph{Experiment Setting}
We conduct experiments on D4RL~\citep{fu2020d4rl} MuJoCo locomotion and AntMaze tasks and three OGBench~\citep{park2025ogbench} manipulation tasks. 
We select two representative O2O RL algorithms as baselines: IQL~\citep{kostrikov2021offline} and Cal-QL~\citep{nakamoto2023cal}. 
For each algorithm, we compare against a dedicated O2O plug-in method: OPT~\citep{shin2025online}, which adds an online critic pre-training stage to IQL, and PARS~\citep{kim2025penalizing}, which regularizes OOD actions for Cal-QL.
More experimental details are provided in Appendix~\ref{appx:implementation-details}.

\subsection{Main Performance Results}
\label{exp:main_exp}

\begin{figure}[htbp]
    \centering
    \vspace{-1ex}
    \includegraphics[width=\textwidth]{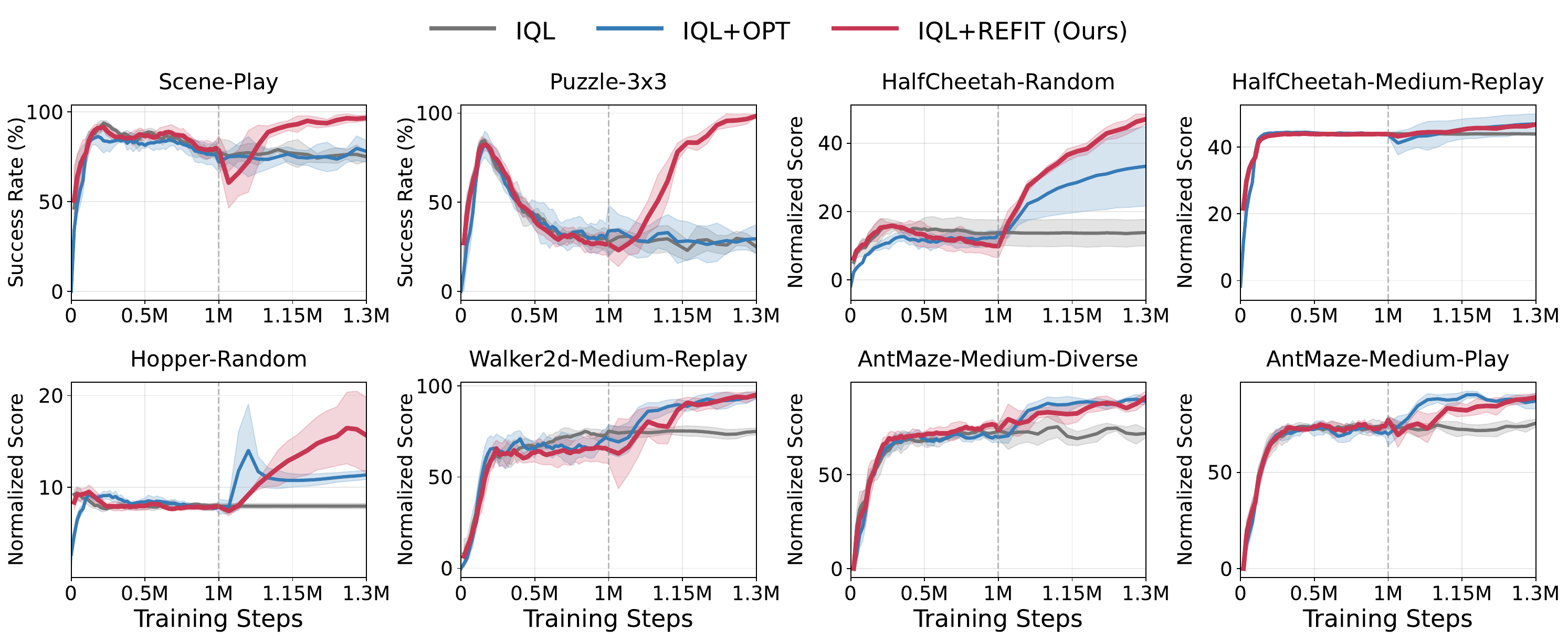}
    \caption{O2O training curves on D4RL and OGBench. The dashed line marks the transition from offline pre-training to online fine-tuning, at which REFIT is applied.
    }
    \label{fig:score_main}
\end{figure}

Following prior work~\citep{shin2025online, kim2025penalizing}, we report performance after 300K online fine-tuning steps. 
Table~\ref{tab:performance_comparison} shows that REFIT improves the aggregate score over vanilla IQL and Cal-QL by 35.4\% and 20.3\% respectively, while also outperforming the corresponding plug-in methods.
The training curves in Figure~\ref{fig:score_main} further show pronounced gains on OGBench and sparse-reward AntMaze tasks, where the base algorithms and even their corresponding plug-in method often plateau during online fine-tuning.
Overall, these results demonstrate that REFIT improves aggregate O2O performance across both backbones, especially in settings where the base algorithms struggle to improve further during online fine-tuning. Additional comparisons are provided in Appendix~\ref{appx:cpr_comparison}.

\vspace{-2ex}

\subsection{Plasticity Diagnostics}
\label{exp:plasticity_exp}

\begin{figure}[H]
    \vspace{-2ex}
    \centering
    \includegraphics[width=\textwidth]{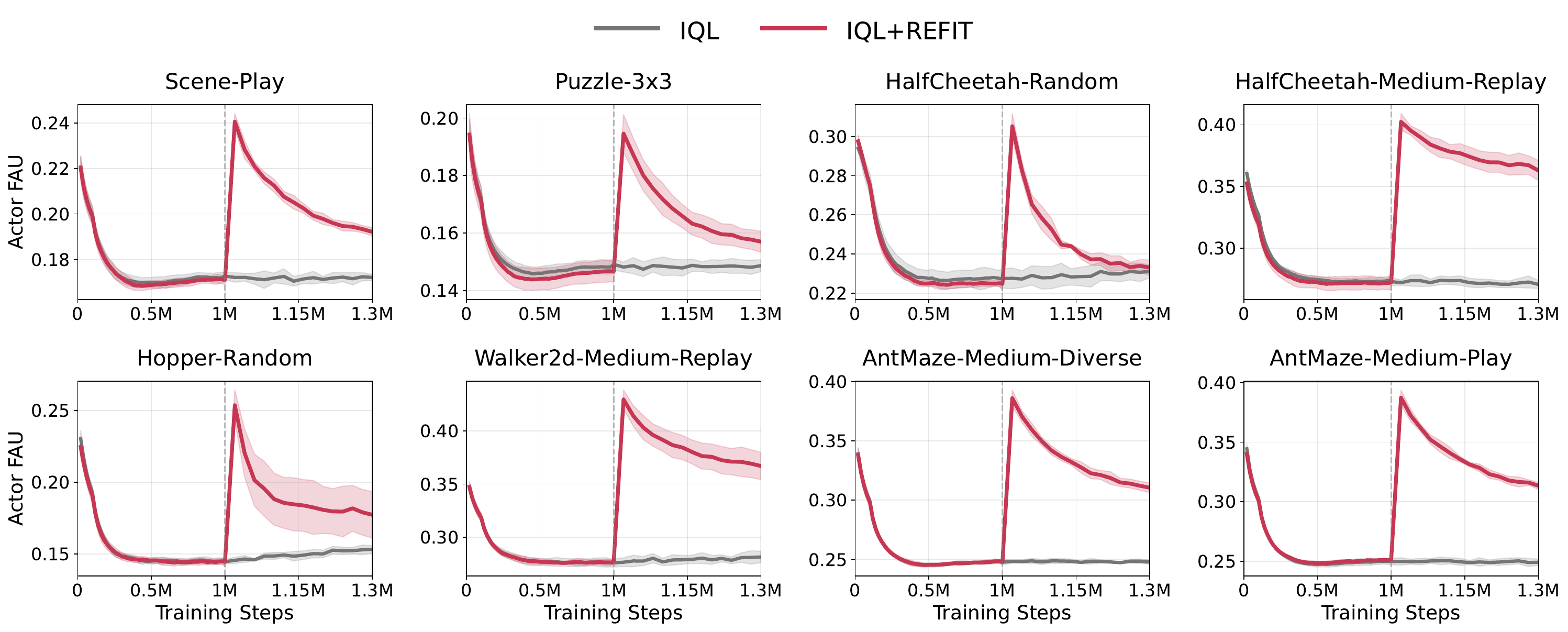}
    \caption{Evolution of actor FAU across the offline-to-online transition.}
    \label{fig:fau_main}
    \vspace{-1ex}
\end{figure}

Figure~\ref{fig:fau_main} tracks the FAU throughout offline and online training. 
The results show that FAU declines during offline pre-training for both methods and remains low under vanilla online fine-tuning. 
In contrast, REFIT sharply restores FAU at the O2O transition, consistent with its goal of restoring network plasticity for subsequent online adaptation.
Meanwhile, Appendix~\ref{appendix:plasticity_analysis} further shows consistent trends across complementary plasticity metrics, including weight norm and SRank, providing additional support for the plasticity-restoring effect of REFIT.

\subsection{Ablation Analysis}
\label{exp:ablation_exp}

\paragraph{Ablation of REFIT Design Components}

We investigate the contributions of fresh initialization, distillation, and partial freezing. 
The \textit{Baseline} directly deploys the classic IQL algorithm, while \textit{Reset} reinitializes the policy network before online fine-tuning.
\textit{Reset + Distillation} further transfers the behavior learned by the offline teacher policy to the freshly initialized student through distillation, without partial freezing. 
Finally, \textit{REFIT} combines all three components.
As shown in Table~\ref{tab:factorial_ablation}, none of these isolated variants can match the performance of the complete REFIT configuration, which achieves the highest overall score.
These results suggest that fresh initialization, policy distillation, and partial freezing provide complementary benefits for subsequent online adaptation.

\begin{table}[htbp]
    \centering
    \caption{Ablation of REFIT design components. All results are averaged over 5 random seeds.}
    \label{tab:factorial_ablation}
    \small
    \setlength{\tabcolsep}{1.5pt}
    \resizebox{\linewidth}{!}{%
    \begin{tabular}{l | cccccc | c}
        \toprule
        Method & \texttt{hc-m} & \texttt{hc-r} & \texttt{wal-m-r} & \texttt{hop-m-r} & \texttt{ant-m-p} & \texttt{ant-m-d} & Score \\
        \midrule
        Baseline             & 48.1 $\pm$ 0.1 & 13.8 $\pm$ 3.9 & 74.8 $\pm$ 2.8 & 77.5 $\pm$ 18.1 & 76.2 $\pm$ 1.5 & 71.6 $\pm$ 2.7 & 362.0 (-23.3\%) \\
        Reset                & 47.3 $\pm$ 0.4 & 38.6 $\pm$ 2.4 & 84.0 $\pm$ 10.6 & 86.5 $\pm$ 29.9 & 71.2 $\pm$ 6.8 & 74.0 $\pm$ 5.5 & 401.6 (-14.9\%) \\
        Reset + Distillation & \underline{49.8 $\pm$ 0.2} & \underline{45.7 $\pm$ 1.9} & \underline{89.3 $\pm$ 11.2} & \underline{97.9 $\pm$ 6.2} & \underline{83.5 $\pm$ 1.8} & \underline{79.2 $\pm$ 23.6} & \underline{445.4} (-5.6\%) \\
        \midrule
        REFIT                & \textbf{50.0} $\pm$ 0.3 & \textbf{46.3} $\pm$ 1.4 & \textbf{96.9} $\pm$ 2.2 & \textbf{102.7} $\pm$ 0.4 & \textbf{86.6} $\pm$ 4.8 & \textbf{89.2} $\pm$ 2.7 & \textbf{471.7} \\
        \bottomrule
    \end{tabular}%
    }
\end{table}

\paragraph{Ablation of Freeze Ratio}

We vary the freeze ratio over $\rho \in \{0.0, 0.25, 0.5, 0.75, 1.0\}$, where $\rho=0.0$ corresponds to distillation without freezing and $\rho=1.0$ freezes the entire student network during distillation. 
As shown in Table~\ref{tab:freeze_ratio_performance}, intermediate freeze ratios achieve comparable aggregate performance, whereas both extreme settings perform notably worse. 
When $\rho=0$, all student parameters are updated during distillation, which may reduce the fraction of active units and limit the plasticity gains from fresh initialization.
Conversely, when $\rho=1$, the student is entirely frozen and retains only the initialized parameters without inheriting the offline behavioral prior. 
The robustness of REFIT across the intermediate range suggests that the core mechanism driving performance is the \emph{combination} of partial freezing and distillation itself, rather than any particular freeze ratio: as long as a sufficient subset of parameters is frozen to preserve plasticity while the remaining parameters absorb the teacher distribution, REFIT maintains comparable performance across these intermediate freeze ratios.
Given its highest aggregate performance across the evaluated tasks, we adopt $\rho=0.5$ as the default in the remaining experiments.

\begin{table}[htbp]
\centering
\caption{
Performance of REFIT under various freeze ratio settings. The two lowest scores in each row are marked in \textcolor{darkred}{red}.}
\label{tab:freeze_ratio_performance}
\small
\setlength{\tabcolsep}{6pt}
\begin{tabular}{lccccc}
\toprule
\textbf{Env} & \textbf{0.0} & \textbf{0.25} & \textbf{0.5} & \textbf{0.75} & \textbf{1.0} \\
\midrule
halfcheetah-m  & \textcolor{darkred}{49.8 $\pm$ 0.2} & 50.0 $\pm$ 0.2 & 50.0 $\pm$ 0.3 & \textbf{50.1} $\pm$ 0.4 & \textcolor{darkred}{47.3 $\pm$ 0.4} \\
halfcheetah-r  & \textcolor{darkred}{45.7 $\pm$ 1.9} & 46.2 $\pm$ 1.3 & 46.3 $\pm$ 1.4 & \textbf{46.5} $\pm$ 1.8 & \textcolor{darkred}{38.6 $\pm$ 2.4} \\
antmaze-m-p    & \textcolor{darkred}{83.5 $\pm$ 1.8} & 86.0 $\pm$ 3.7 & \textbf{86.6} $\pm$ 4.8 & 85.8 $\pm$ 4.1 & \textcolor{darkred}{71.2 $\pm$ 6.8} \\
antmaze-m-d    & \textcolor{darkred}{79.2 $\pm$ 23.6} & 86.4 $\pm$ 3.0 & \textbf{89.2} $\pm$ 2.7 & 84.4 $\pm$ 1.5 & \textcolor{darkred}{74.0 $\pm$ 5.5} \\
\midrule
\rowcolor[gray]{0.9} \textbf{Total} & \textcolor{darkred}{258.2} & 268.6 & \textbf{272.1} & 266.8 & \textcolor{darkred}{231.1} \\
\bottomrule
\end{tabular}
\end{table}

\paragraph{Ablation of Distillation Target}
\label{ablation:distillation_comp}

\vspace{0ex}
\begin{figure}[H]
    \centering
    \includegraphics[width=\textwidth]{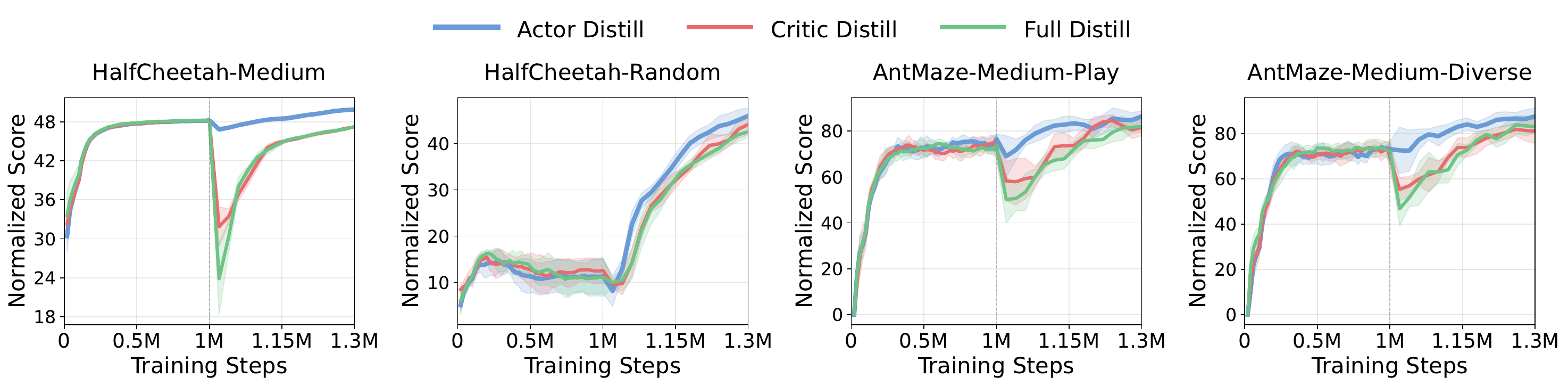}
    \caption{Learning curves of REFIT with different distillation targets.}
    \label{fig:distill_methods}
\end{figure}

We compare three distillation targets: actor-only, critic-only, and full actor-critic distillation. 
As shown in Figure~\ref{fig:distill_methods}, variants involving critic distillation exhibit a pronounced performance drop at the start of online fine-tuning, whereas actor-only distillation largely avoids this degradation and maintains more stable adaptation. 
This suggests that critic distillation introduces additional perturbations to the learned value function, making the O2O transition less stable.
Based on this observation, REFIT distills only the actor network while allowing the critic to adapt freely during online fine-tuning.

\vspace{1ex}
\begin{wrapfigure}{r}{0.46\textwidth}
  \centering
  \vspace{0ex}
  \includegraphics[width=\linewidth]{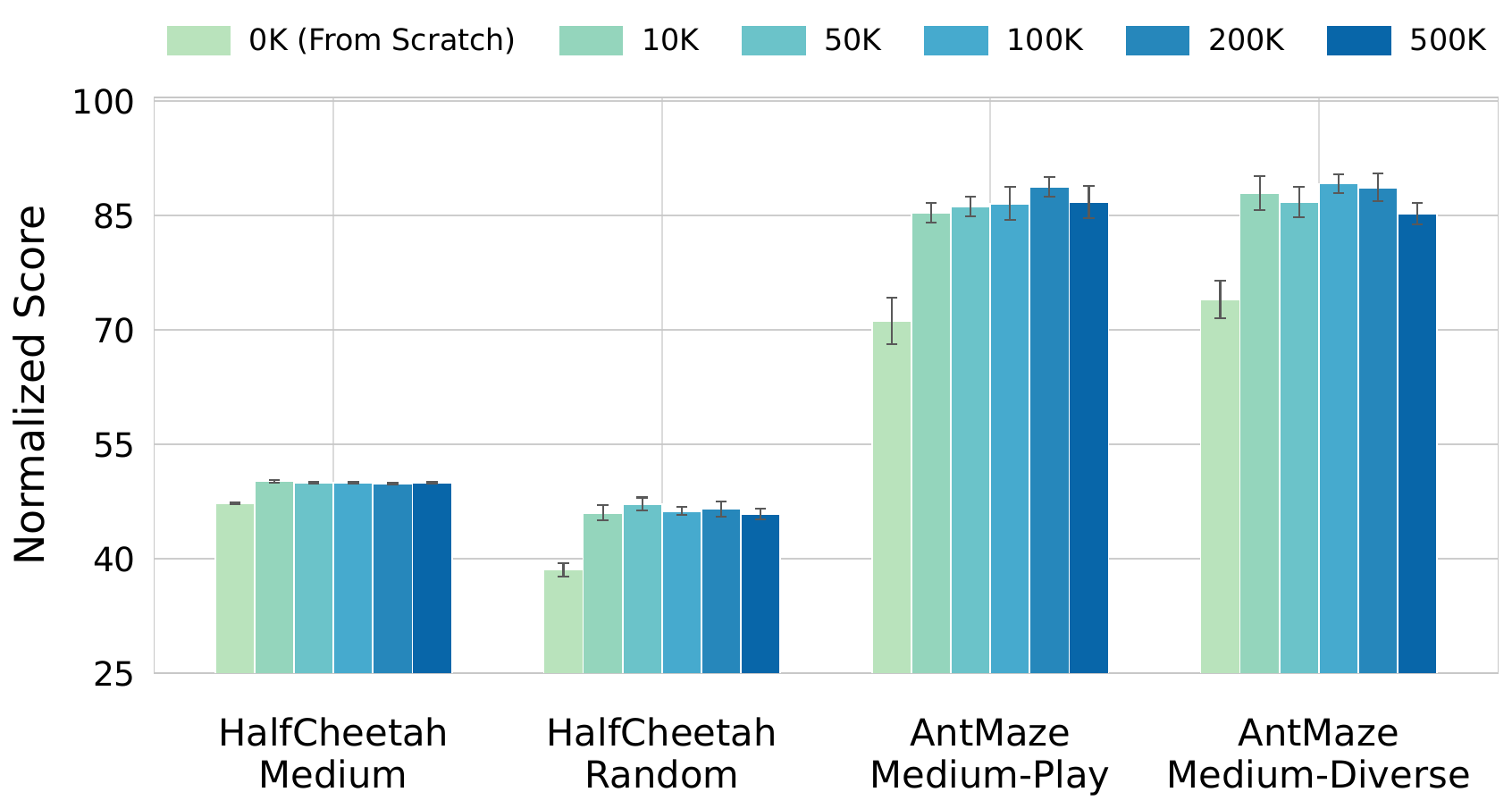}
  \vspace{-3ex}
  \caption{Performance of REFIT under different distillation budget settings.}
  \label{fig:distill_steps}
  \vspace{-2ex}
\end{wrapfigure}

\paragraph{Ablation of Distillation Budget}
We vary the policy-distillation budget from 0K, corresponding to no distillation, to 500K updates. As shown in Figure~\ref{fig:distill_steps}, all nonzero distillation budgets consistently outperform the 0K baseline, highlighting the importance of transferring the offline policy before online fine-tuning. 
Meanwhile, performance remains relatively stable across various settings, indicating limited sensitivity to the exact distillation duration. 
Together with the freeze ratio ablation, REFIT remains effective across a wide range of distillation configurations without requiring careful hyperparameter tuning.
Based on this robustness, we adopt 100K updates as the default setting, which provides strong performance without incurring unnecessary additional computation.

\subsection{Robustness to Offline Training Budget}
\label{exp:offline_budget}

\begin{figure}[htbp]
    \centering
    \begin{subfigure}[t]{0.32\textwidth}
        \centering
        \includegraphics[width=\textwidth]{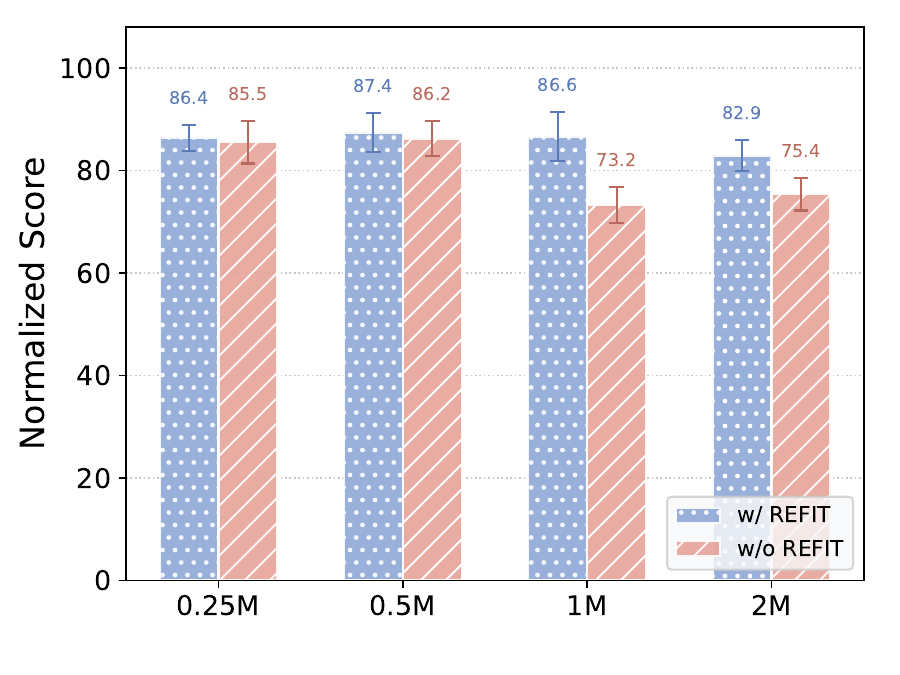}
        \caption{Online performance}
        \label{fig:offline_budget_a}
    \end{subfigure}%
    \hfill
    \begin{subfigure}[t]{0.32\textwidth}
        \centering
        \includegraphics[width=\textwidth]{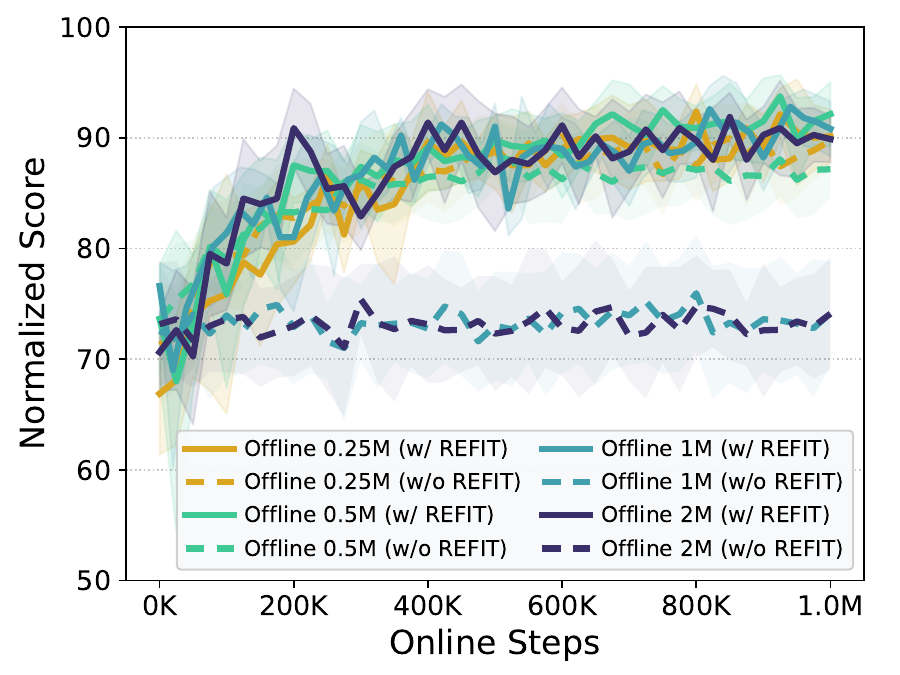}
        \caption{Online learning curves}
        \label{fig:offline_budget_b}
    \end{subfigure}%
    \hfill
    \begin{subfigure}[t]{0.32\textwidth}
        \centering
        \includegraphics[width=\textwidth]{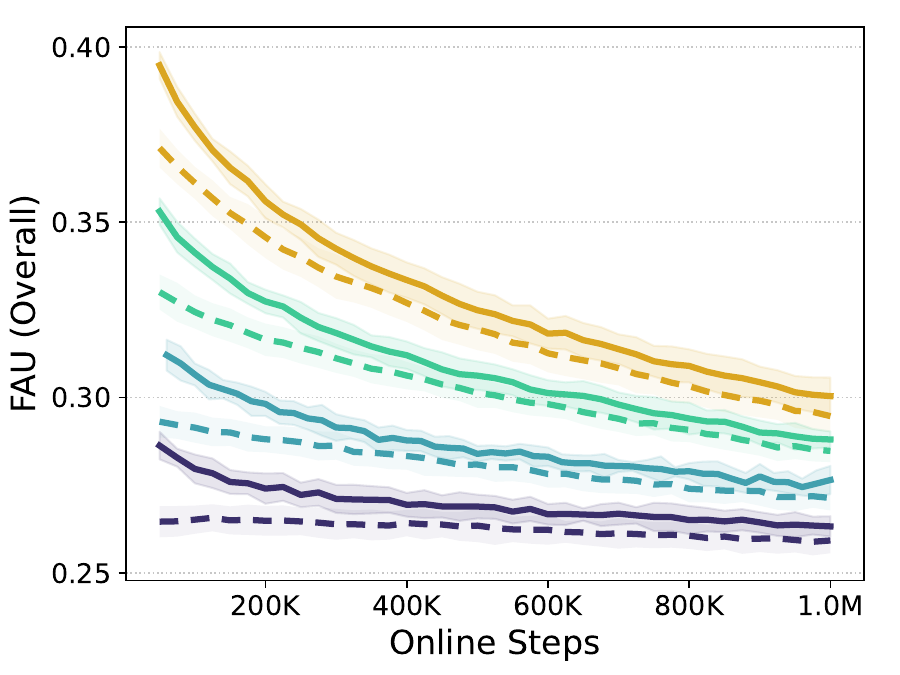}
        \caption{FAU}
        \label{fig:offline_budget_c}
    \end{subfigure}
    \caption{
    Robustness of REFIT across different offline training budgets.
    (a)~Normalized online performance of REFIT and the corresponding IQL baseline after online fine-tuning under different offline training budgets.
    (b)~Online learning curves of REFIT and the baseline across different offline training budgets.
    (c)~FAU during online fine-tuning.
    }
    \label{fig:offline_budget_robustness}
\end{figure}

The motivation study in Section~\ref{plasticity_toy} shows that prolonged offline optimization can degrade network plasticity and hinder subsequent online adaptation.
We therefore evaluate REFIT across offline training budgets ranging from $0.25\text{M}$ to $2\text{M}$ steps under the same setting as in Section~\ref{plasticity_toy}. 
As shown in Figure~\ref{fig:offline_budget_a} and Figure~\ref{fig:offline_budget_b}, REFIT consistently outperforms the corresponding baseline across all offline budget settings. 
Notably, as the offline training budget increases, the baseline exhibits increasingly pronounced plateaus during online fine-tuning, whereas REFIT continues to improve, suggesting that restoring network plasticity enables effective adaptation after prolonged offline optimization. 
Figure~\ref{fig:offline_budget_c} further shows that REFIT maintains higher FAU throughout online fine-tuning across all offline budget settings, providing further evidence of improved network plasticity. 
Together, these results demonstrate the effectiveness of REFIT in mitigating the plasticity degradation induced by prolonged offline optimization and facilitating subsequent online adaptation.

\vspace{-2ex}
\section{Conclusion}

In this work, we investigate O2O RL from the perspective of network plasticity. 
Our empirical analysis shows that prolonged optimization on static offline data can progressively reduce network plasticity even after offline performance has largely saturated, and that policies retaining higher plasticity tend to exhibit stronger subsequent online adaptation. 
Motivated by these observations, we propose REFIT, a lightweight model-level method that distills the behavior of an offline teacher into a freshly initialized student policy while temporarily freezing a random subset of student units.
This procedure preserves the offline behavioral prior while restoring policy plasticity for subsequent online adaptation. 
Experiments across D4RL and OGBench demonstrate that REFIT improves aggregate performance for IQL and Cal-QL and achieves higher aggregate scores than the corresponding O2O plug-in methods. 
These results highlight policy plasticity as an important factor in O2O adaptation and demonstrate the effectiveness of explicitly restoring plasticity at the O2O transition.

\subsection*{AI use statement}
In this paper, we have used generative AI tools for implementing part of the code of our method. We have not used generative AI tools for helping develop theoretical models or conceptual frameworks, formulating mathematical claims, providing critical ingredients for proving mathematical claims and interpreting results. The rest of the required disclosure tasks are not applicable to this work. Additionally, we used generative AI tools to refine the expression of the paper. All LLM-generated content has been reviewed by the authors. We take responsibility for the final content of this work, including text, claims or artifacts produced with the aid of
generative AI.

\subsection*{Ethics statement}
This work studies plasticity loss in O2O RL and proposes a model-level method for improving policy fine-tuning. All experiments are conducted in simulated continuous-control, navigation, and robotic-manipulation benchmarks, including D4RL and OGBench, and do not involve human subjects, personal data or safety-critical systems. Our method does not introduce capabilities that directly target surveillance, identity inference, autonomous weapons, or other clearly harmful applications. While improved O2O RL could in principle accelerate the deployment of learned policies in downstream systems, we do not identify ethical concerns specific to this work beyond those common to general RL research. We therefore do not believe this work requires additional mitigation protocols.

\subsection*{Reproducibility statement}
We have made efforts to ensure the reproducibility of our work.
To facilitate reproducibility, we provide our code, including hyperparameter configuration files in the supplementary material, with implementation details, hyperparameters, and hardware specifications documented in Appendix~\ref{appx:implementation-details}.

\bibliography{references}
\bibliographystyle{iclr2027_conference}
\clearpage

\appendix

\section{Pseudocode}
\label{appx:REFIT-algorithm}

\begin{algorithm}[H]
\caption{REFIT Training Procedure}
\label{alg:REFIT-overall}
\begin{algorithmic}[1]
\REQUIRE Offline dataset $\mathcal{D}_{\text{off}}$; environment $E$; O2O algorithm; freeze ratio $\rho$.
\REQUIRE Budgets $N_{\text{off}}=10^6$, $N_{\text{distill}}=10^5$, and $N_{\text{on}}=3\times10^5$.
\STATE \textbf{Phase 1: Offline Pre-training}
\FOR{$t = 1$ to $N_{\text{off}}$}
    \STATE Sample batch $(s, a, r, s') \sim \mathcal{D}_{\text{off}}$.
    \STATE Update policy $\pi_\theta$ and backbone value functions using the offline RL objective.
\ENDFOR
\STATE \textbf{Phase 2: Plasticity-Preserving Policy Distillation}
\STATE Set $\pi_{\text{teacher}} \leftarrow \pi_\theta$ and initialize a fresh student policy $\pi_{\text{student}}$.
\STATE Randomly partition $\theta_{\text{student}}$ into frozen parameters $\theta_{\text{freeze}}$ and trainable parameters $\theta_{\text{train}}$ according to the freeze ratio $\rho$.
\FOR{$k = 1$ to $N_{\text{distill}}$}
    \STATE Sample $s \sim \mathcal{D}_{\text{off}}$ and update unfrozen student units using $\mathcal{L}_{\text{distill}}$ in Eq.~(\ref{eq:distill_formula})
\ENDFOR
\STATE Replace $\pi_\theta \leftarrow \pi_{\text{student}}$; retain the offline-trained value functions.
\STATE \textbf{Phase 3: Online Fine-tuning}
\STATE Unfreeze $\theta_{\text{freeze}}$, making all student policy parameters trainable.
\FOR{$t = 1$ to $N_{\text{on}}$}
    \STATE Interact with $E$, add the transition to the replay buffer, update with online RL objective.
\ENDFOR
\end{algorithmic}
\end{algorithm}

\section{Experiment Details}
\label{appx:implementation-details}

\subsection{Environment Details}
\label{appx:env-dataset}

We evaluate D4RL MuJoCo and AntMaze~\citep{fu2020d4rl} and three OGBench manipulation tasks~\citep{park2025ogbench}. Table~\ref{tab:environment-names} defines the compositional abbreviations used in our result tables. To fit space-constrained tables, we additionally shorten the environment prefixes to \texttt{hc}, \texttt{hop}, \texttt{wal}, and \texttt{ant}; for example, \texttt{halfcheetah-m}, \texttt{walker2d-m-r}, and \texttt{antmaze-m-p} become \texttt{hc-m}, \texttt{wal-m-r}, and \texttt{ant-m-p}, respectively. OGBench task names are not abbreviated.

\begin{table}[H]
\caption{Environment naming convention. Tokens are composed from left to right, as illustrated in the examples.}
\label{tab:environment-names}
\centering
\small
\setlength{\tabcolsep}{5pt}
\begin{tabular}{llll}
\toprule
\textbf{Suite} & \textbf{Token} & \textbf{Meaning} & \textbf{Example} \\
\midrule
\multirow{4}{*}{Environment}
& \texttt{hc} & \texttt{halfcheetah} & \texttt{hc-m} \\
& \texttt{hop} & \texttt{hopper} & \texttt{hop-m-r} \\
& \texttt{wal} & \texttt{walker2d} & \texttt{wal-r} \\
& \texttt{ant} & \texttt{antmaze} & \texttt{ant-l-d} \\
\midrule
\multirow{3}{*}{MuJoCo Postfix}
& \texttt{r} & \texttt{random} & \texttt{hc-r} \\
& \texttt{m} & \texttt{medium} & \texttt{hop-m} \\
& \texttt{m-r} & \texttt{medium-replay} & \texttt{wal-m-r} \\
\midrule
\multirow{4}{*}{AntMaze Postfix}
& \texttt{m-d} & \texttt{medium-diverse} & \texttt{ant-m-d} \\
& \texttt{m-p} & \texttt{medium-play} & \texttt{ant-m-p} \\
& \texttt{l-d} & \texttt{large-diverse} & \texttt{ant-l-d} \\
& \texttt{l-p} & \texttt{large-play} & \texttt{ant-l-p} \\
\bottomrule
\end{tabular}
\end{table}

\subsection{REFIT Implementation}
\label{appx:REFIT-implementation}

All experiments use NVIDIA A100 GPUs with CUDA 12.4.

\paragraph{Three-phase training.}
\label{appx:procedure}
We first train the selected backbone on $\mathcal{D}_{\text{off}}$ for $10^6$ steps. At the O2O transition, we distill the offline teacher into a freshly initialized student policy for $10^5$ steps while applying a fixed mask that freezes the kernel columns and biases of \(\lfloor\rho d_{\mathrm{out}}\rfloor\) randomly selected output channels in each actor Dense layer, including the policy output head; the critic and value networks are inherited from the offline phase. We then remove the mask and fine-tune all networks for $3\times10^5$ online steps. Algorithm~\ref{alg:REFIT-overall} summarizes the full procedure.

\paragraph{Hyperparameters.}
\label{appx:hyperparams}

Across all environments and both backbones, we use $10^6$ offline training steps, $10^5$ policy-distillation updates, and $3\times10^5$ online fine-tuning steps. The default freeze ratio is $\rho=0.5$; for Cal-QL on MuJoCo, we use $\rho=0.1$ for HalfCheetah and Hopper and $\rho=0.5$ for Walker2d. Other settings, such as network architecture, policy learning rate, IQL expectile and temperature, and Cal-QL conservative weight, follow the corresponding backbone implementations. Complete environment-specific configurations, including batch size and normalization, are provided in the configuration files of the supplementary code.

\vspace{2ex}
\subsection{Backbone Implementation}
\label{appx:backbone}
\vspace{1ex}

We evaluate REFIT on IQL~\citep{kostrikov2021offline} and Cal-QL~\citep{nakamoto2023cal}, comparing against OPT~\citep{shin2025online} and PARS~\citep{kim2025penalizing}, respectively. Our implementations build on the following public repositories:
\begin{itemize}[nosep,leftmargin=*]
    \item IQL: ~\url{https://github.com/ikostrikov/implicit_q_learning}
    \item Cal-QL: ~\url{https://github.com/nakamotoo/Cal-QL}
    \item OPT: ~\url{https://github.com/LGAI-Research/opt}
    \item PARS: ~\url{https://github.com/LGAI-Research/pars}
\end{itemize}

\section{Computational Cost}
\label{appendix:time}

In this section, we evaluate the computational overhead of REFIT by comparing the wall-clock time of IQL+REFIT with that of the IQL baseline on \texttt{halfcheetah-random-v2}. All runs are conducted on a single NVIDIA A100 GPU and averaged over five random seeds.

\vspace{2ex}
\begin{figure}[htbp]
    \centering
    \includegraphics[width=0.6\linewidth]{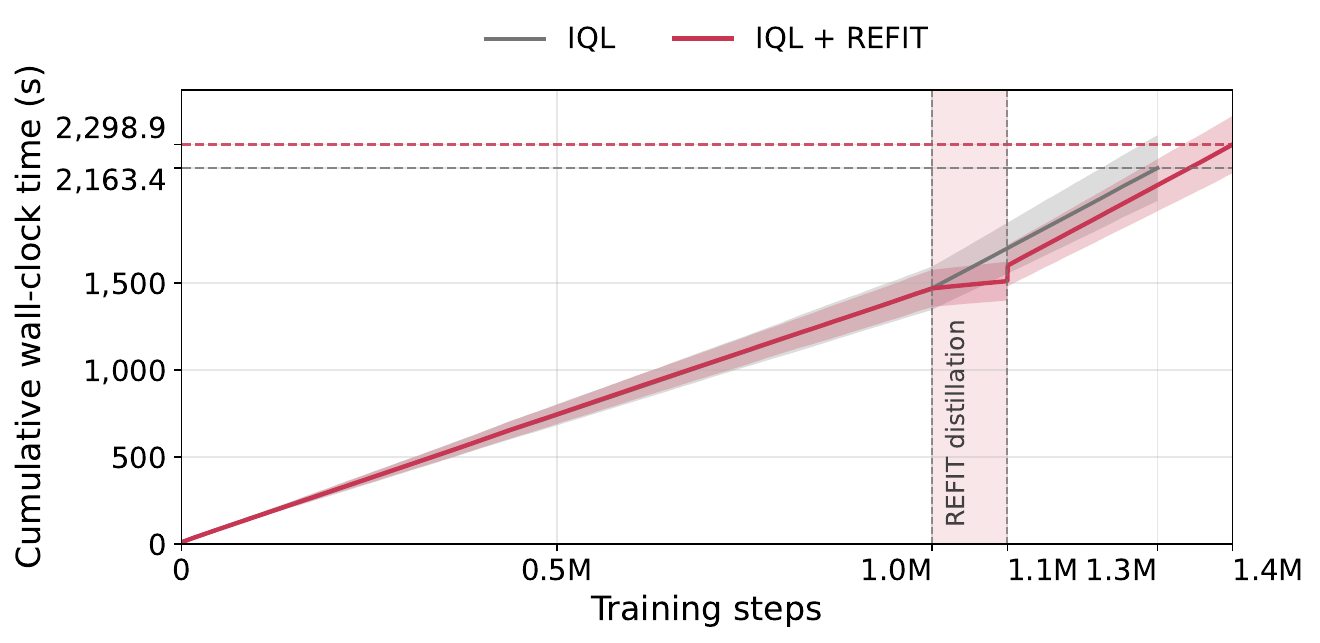}
    \caption{Wall-clock time of IQL+REFIT and the IQL baseline, averaged over five random seeds.}
    \label{fig:computational_cost}
\end{figure}
\vspace{2ex}

As shown in Figure~\ref{fig:computational_cost}, the complete pipeline requires 2163.4 seconds for IQL and 2298.9 seconds for IQL+REFIT, corresponding to a $6.26\%$ increase in wall-clock time. The small offset near the end of distillation arises from one-time JAX just-in-time compilation when the distillation update and student policy are initialized. After compilation, the two curves progress approximately in parallel, indicating that REFIT introduces limited sustained overhead beyond the additional distillation phase.

\section{Additional Comparison with CPR}
\label{appx:cpr_comparison}

CPR~\citep{kong2024efficient} links primacy bias to limited O2O adaptation, but does not directly quantify offline-induced policy plasticity loss with network-level diagnostics. Its periodic policy revitalization retains earlier policies for action selection. REFIT instead diagnoses this loss and applies a single model-level intervention at the O2O transition, then fine-tunes one actor online. Table~\ref{tab:cpr_comparison} compares both methods on IQL and Cal-QL across six D4RL tasks. We adapt CPR from their released code (\url{https://github.com/LAMDA-RL/CPR}), following its method-specific hyperparameters where applicable.

\vspace{1ex}
\begin{table}[htbp]
    \centering
    \caption{Normalized scores on six representative tasks. }
    \label{tab:cpr_comparison}
    \small
    \setlength{\tabcolsep}{3pt}
    \resizebox{\textwidth}{!}{%
    \begin{tabular}{lcccccc|c}
        \toprule
        Method & \texttt{hc-m} & \texttt{hc-r} & \texttt{wal-m-r} & \texttt{hop-m-r} & \texttt{ant-m-p} & \texttt{ant-m-d} & Total \\
        \midrule
        IQL & 48.1 $\pm$ 0.1 & 13.8 $\pm$ 3.9 & 74.8 $\pm$ 2.8 & 77.5 $\pm$ 18.1 & \underline{76.2} $\pm$ 1.5 & \underline{71.6} $\pm$ 2.7 & 362.0 \\
        IQL+CPR & \textbf{51.7} $\pm$ 0.5 & \underline{26.0} $\pm$ 4.0 & \underline{86.0} $\pm$ 11.7 & \underline{101.8} $\pm$ 3.9 & 66.0 $\pm$ 3.3 & 71.0 $\pm$ 6.8 & \underline{402.5} \\
        IQL+REFIT & \underline{50.0} $\pm$ 0.3 & \textbf{46.3} $\pm$ 1.4 & \textbf{96.9} $\pm$ 2.2 & \textbf{102.7} $\pm$ 0.4 & \textbf{86.6} $\pm$ 4.8 & \textbf{89.2} $\pm$ 2.7 & \textbf{471.7} \\
        \midrule
        Cal-QL & \textbf{65.3} $\pm$ 3.0 & \underline{33.4} $\pm$ 3.8 & \underline{86.7} $\pm$ 8.7 & \underline{68.9} $\pm$ 7.8 & \underline{92.2} $\pm$ 2.1 & \underline{93.2} $\pm$ 0.8 & \underline{439.7} \\
        Cal-QL+CPR & 43.3 $\pm$ 0.6 & 2.2 $\pm$ 0.1 & 17.6 $\pm$ 2.8 & 30.3 $\pm$ 7.9 & 27.0 $\pm$ 10.3 & 26.0 $\pm$ 13.9 & 146.3 \\
        Cal-QL+REFIT & \underline{51.1} $\pm$ 0.9 & \textbf{39.2} $\pm$ 3.5 & \textbf{97.9} $\pm$ 5.8 & \textbf{87.6} $\pm$ 7.5 & \textbf{95.4} $\pm$ 1.7 & \textbf{96.2} $\pm$ 0.8 & \textbf{467.4} \\
        \bottomrule
    \end{tabular}%
    }
\end{table}
\vspace{1ex}

These results demonstrate that REFIT outperforms CPR in most cases by directly restoring policy plasticity at the model level.

\vspace{1ex}
\section{Supplementary Ablation Analysis}
\vspace{1ex}
\label{appendix:ablation_exp}

This section provides supplementary ablations that complement the analysis in Section~\ref{exp:ablation_exp}. We examine the necessity of policy distillation and network reinitialization, the sensitivity to the distillation loss function, the effect of standard-deviation transfer on exploration, and the role of unfreezing during online adaptation.

\vspace{4ex}
\subsection{Ablation of Policy Distillation}
\label{appendix:necessity_distillation}
To evaluate the role of distillation in mitigating policy degradation during the O2O transition, we compare REFIT against a \textit{direct reset} baseline (referred to as From-Scratch), which randomly reinitializes the network parameters at the start of online fine-tuning without any behavioral prior from the offline teacher.

\vspace{2ex}
\begin{figure}[htbp]
    \centering
    \includegraphics[width=\textwidth]{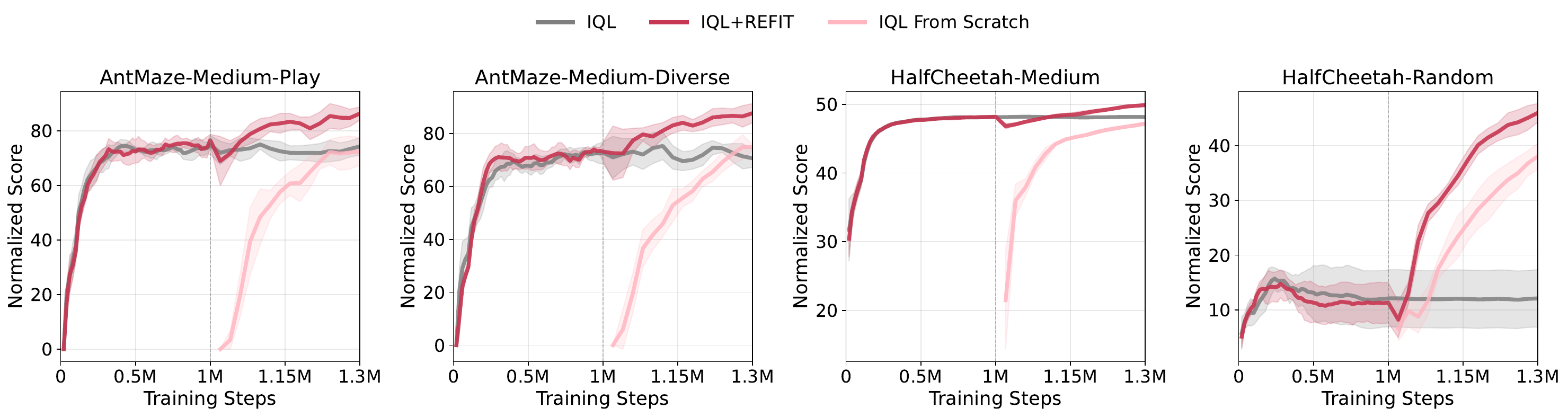}
    \caption{Performance comparison between IQL, IQL From Scratch and IQL+REFIT}
    \label{fig:from_scratch}
\end{figure}
\vspace{2ex}

As shown in Figure~\ref{fig:from_scratch}, although the direct reinitialization baseline outperforms standard fine-tuning on aggregate by restoring network plasticity, it consistently falls short of the full REFIT configuration. This result confirms that combining policy distillation with structured reinitialization effectively preserves the offline behavioral prior, whereas naive parameter resetting discards the behavioral prior acquired during offline training.

\vspace{2ex}
\subsection{Ablation of Policy Reinitialization}
\label{appendix:necessity_reset}

\begin{wrapfigure}{r}{0.36\textwidth}
  \centering
  \vspace{-10pt}
  \includegraphics[width=\linewidth]{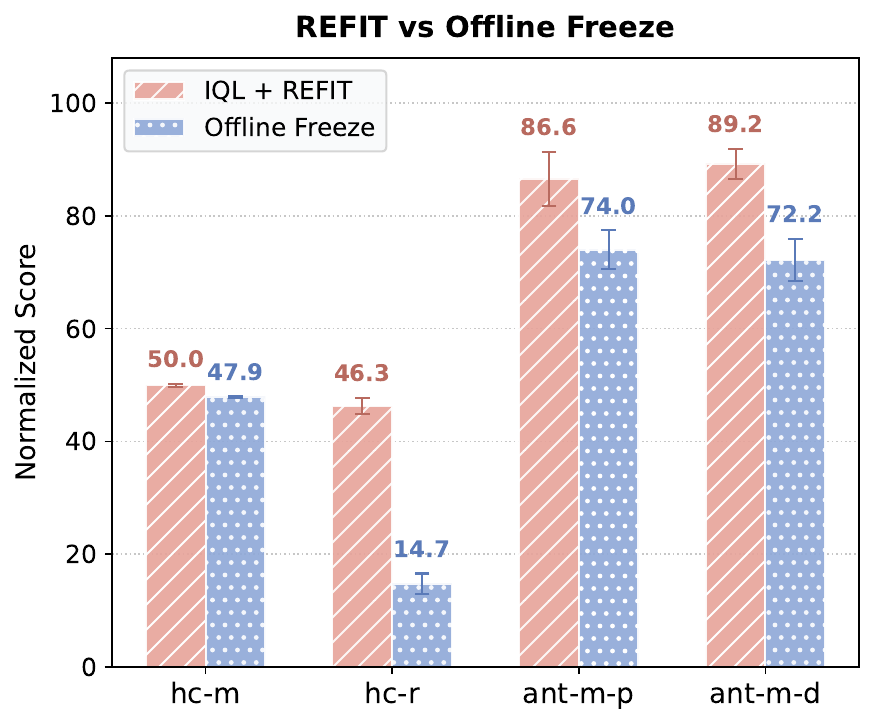}
  \caption{Impact of network reset during distillation.}
  \label{fig:necessity_of_reset}
  \vspace{-10pt}
\end{wrapfigure}

A natural question arises regarding the distillation phase: \textit{is it essential to freshly initialize the online student policy}, or could one simply freeze parts of the network during offline pre-training and unfreeze them for online fine-tuning? To isolate the effect of network reinitialization, we compare the default REFIT method against this unfreezing baseline across four representative environments. As shown in Figure~\ref{fig:necessity_of_reset}, directly unfreezing offline pre-trained parameters leads to substantial performance degradation, particularly on tasks requiring significant adaptation (e.g., \texttt{halfcheetah-r} drops from 46.3 to 14.7). These results indicate that fresh initialization is essential for overcoming the optimization bottlenecks accumulated during offline training and enabling effective online policy improvement.

\subsection{Ablation of Distillation Loss Function}
\label{appendix:distill_loss}

\begin{wraptable}{r}{0.54\textwidth}
\vspace{-12pt}
\centering
\caption{Ablation of distillation loss. Results report mean $\pm$ standard deviation.}
\label{tab:distill_loss_ablation}
\small
\setlength{\tabcolsep}{2.5pt}
\resizebox{0.54\textwidth}{!}{%
\begin{tabular}{lccc}
\toprule
\textbf{Environment} & \textbf{$\text{KL}_{\text{fwd}}$} & \textbf{$\text{KL}_{\text{rev}}$} & \textbf{MSE} \\
\midrule
antmaze-m-d   & 86.9 $\pm$ 6.7          & 87.0 $\pm$ 5.2          & \textbf{89.2} $\pm$ 2.7 \\
antmaze-m-p   & \textbf{87.8} $\pm$ 2.7 & 84.8 $\pm$ 5.2          & 86.6 $\pm$ 4.8          \\
halfcheetah-m & 49.9 $\pm$ 0.2          & 50.0 $\pm$ 0.6          & 50.0 $\pm$ 0.3          \\
halfcheetah-r & 46.3 $\pm$ 1.5          & \textbf{47.5} $\pm$ 1.7 & 46.3 $\pm$ 1.4          \\
\midrule
\textbf{Total Score} & 270.9 $\pm$ 2.9  & 269.2 $\pm$ 2.9         & \textbf{272.1} $\pm$ 2.8 \\
\bottomrule
\end{tabular}%
}
\vspace{-8pt}
\end{wraptable}
REFIT adopts mean squared error~(MSE) as its default distillation objective due to computational simplicity: MSE directly regresses the student policy toward the mean action of the frozen teacher policy, without requiring action sampling or full distribution matching. To verify that the method is not tied to a particular choice of loss function, we compare MSE against forward KL divergence and reverse KL divergence, keeping all other components unchanged.

As shown in Table~\ref{tab:distill_loss_ablation}, all three objectives yield comparable aggregate performance across the evaluated environments, with per-task differences largely within the reported standard deviations. This result confirms that the effectiveness of REFIT stems from its structural design---fresh initialization, partial freezing, and knowledge transfer---rather than from the specific form of the distillation loss. We therefore adopt MSE as the default for its simplicity.

\subsection{Ablation of Standard-Deviation Transfer}
\label{appendix:copy_std}

Because REFIT reinitializes the student network, the log-standard-deviation parameters of the policy are also reset to their default initial values, which may differ from those learned during offline training. A natural concern is whether the performance gain of REFIT is primarily attributable to the increased policy standard deviation induced by this reinitialization, rather than to the restoration of network plasticity.

\begin{table}[htbp]
\centering
\caption{Ablation of standard-deviation transfer. ``Default'' denotes the standard REFIT configuration where the log-standard-deviation parameters are reset along with the rest of the student network; ``Copy Std'' denotes the variant that inherits the standard deviation from the offline teacher. Results are reported as mean $\pm$ standard deviation across random seeds.}
\label{tab:copy_std_ablation}
\small
\setlength{\tabcolsep}{4.5pt}
\begin{tabular}{lccccc}
\toprule
\textbf{Method} & \textbf{antmaze-m-d} & \textbf{antmaze-m-p} & \textbf{halfcheetah-m} & \textbf{halfcheetah-r} & \textbf{Total Score} \\
\midrule
Default  & \textbf{89.2 $\pm$ 2.7} & \textbf{86.6 $\pm$ 4.8} & 50.0 $\pm$ 0.3          & 46.3 $\pm$ 1.4          & \textbf{272.1 $\pm$ 2.8} \\
Copy Std & 86.4 $\pm$ 4.5          & 84.3 $\pm$ 4.0          & \textbf{50.1 $\pm$ 0.3} & \textbf{46.9 $\pm$ 2.9} & 267.7 $\pm$ 2.3          \\
\bottomrule
\end{tabular}
\end{table}

To disentangle these two factors, we evaluate a \textit{Copy Std} variant that directly inherits the standard-deviation parameters from the offline teacher after distillation, so that the student begins online fine-tuning with the same exploration profile as the teacher. As shown in Table~\ref{tab:copy_std_ablation}, the Copy Std variant achieves performance comparable to the default REFIT configuration across all evaluated environments, with aggregate scores differing by only a small margin. This result indicates that the performance gain of REFIT does not rely on increased exploration induced by resetting the standard deviation, and is instead consistently attributable to the plasticity restoration provided by the structured reinitialization and distillation procedure.

\subsection{Ablation of Online Unfreezing}
\label{appendix:online_freeze}

\begin{figure}[htbp]
    \centering
    \includegraphics[width=\textwidth]{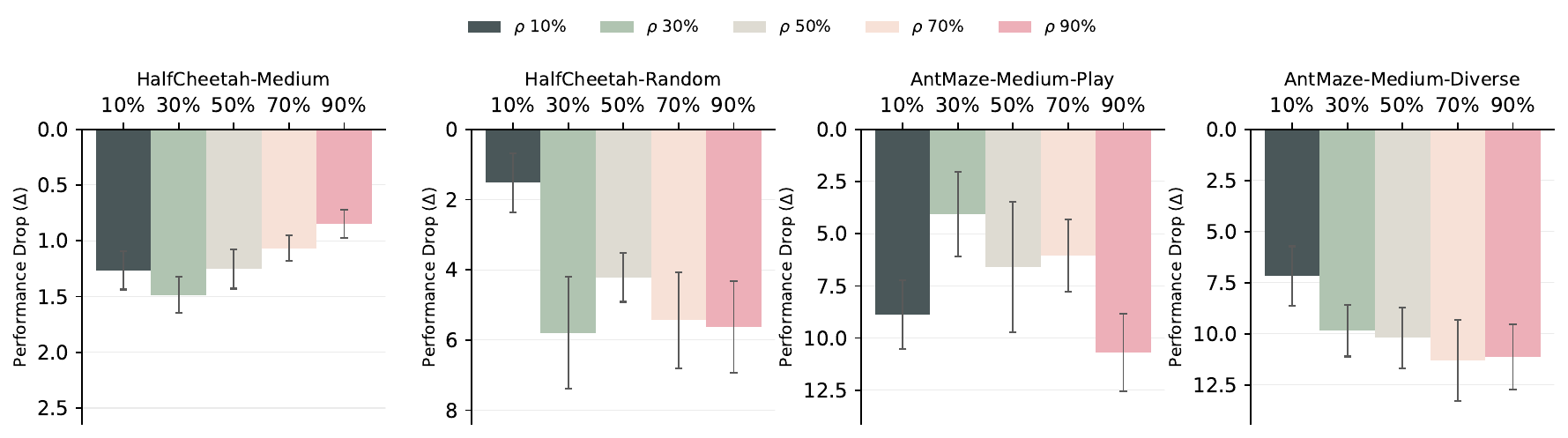}
    \caption{Online performance with and without unfreezing the parameter subset after distillation.}
    \label{fig:online_freeze}
\end{figure}

A final design question is whether the parameter subset frozen during distillation should remain frozen during online fine-tuning. The \textit{online-freeze} variant retains the same mask throughout the online phase, whereas the default REFIT configuration removes the mask and updates the entire policy. As shown in Figure~\ref{fig:online_freeze}, retaining the mask consistently limits online improvement and yields lower asymptotic performance. Unfreezing is therefore necessary to recover the full optimization capacity of the student policy during online adaptation.

\section{Supplementary Plasticity Diagnostics}

\subsection{Definitions of Additional Plasticity Metrics}
\label{appendix:plasticity_metrics}

Here we introduce the two other metrics we used in the experiment: 

1. \textit{Weight Norm}: The magnitude of network parameters provides a complementary diagnostic of network plasticity~\citep{lyle2024disentangling}. We track the L2 norm of all actor parameters:
\begin{equation}
    \|\boldsymbol{\theta}\|_{\text{weight}}
    = \sqrt{\sum_{i=1}^{P}\theta_i^2},
\end{equation}
where $\boldsymbol{\theta}\in\mathbb{R}^{P}$ contains all actor parameters, including weights, biases, and policy standard-deviation parameters. We interpret changes in this norm alongside FAU and SRank, rather than as direct evidence of plasticity loss or restoration.

2. \textit{SRank}: SRank measures the effective dimensionality of learned representations~\citep{kumar2020implicit}. For an actor hidden layer, let $F \in \mathbb{R}^{d \times m}$ denote the matrix of post-ReLU activations for $d$ sampled states and hidden width $m$, and let $\sigma_1(F) \geq \cdots \geq \sigma_q(F)$ be its singular values, where $q=\min(d,m)$. We define SRank as the minimum number of leading singular values required to retain a fraction $1-\epsilon$ of the total singular-value mass:
\begin{equation}
    \operatorname{SRank}_{\epsilon}(F)
    =
    \min\left\{
    k:
    \frac{\sum_{j=1}^{k}\sigma_j(F)}
         {\sum_{j=1}^{q}\sigma_j(F)}
    \geq 1-\epsilon
    \right\},
    \qquad \epsilon=0.01.
\end{equation}
Unlike an absolute singular-value threshold, this relative criterion is invariant to a global rescaling of $F$ and therefore separates changes in spectral shape from changes in feature magnitude. We compute SRank separately for each actor hidden layer using 10 mini-batches of $d=256$ states sampled uniformly from the replay buffer, average the resulting values across batches and layers, and record the metric every 5000 policy updates. Because $d=256$, the reported SRank is upper bounded by 256. A higher value indicates that the actor representation distributes its spectral mass across more directions, providing a complementary measure of representational capacity rather than a direct measure of performance.

\subsection{Additional Diagnostic Results}
\label{appendix:plasticity_analysis}

\paragraph{Weight Norm}
\vspace{1ex}
\begin{figure}[htbp]
    \centering
    \includegraphics[width=\textwidth]{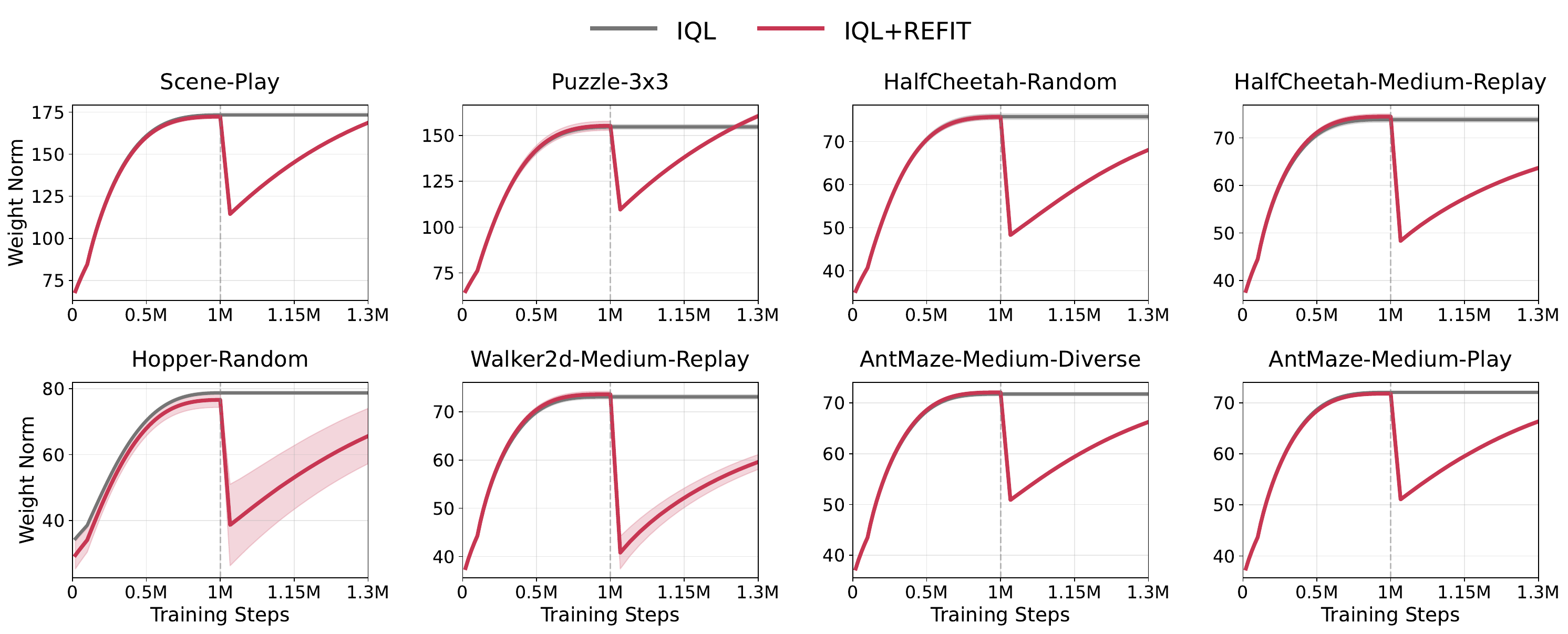}
    \caption{Weight Norm of REFIT and baseline throughout the training process.}
    \label{fig:weight_norm_main}
\end{figure}
\vspace{2ex}

Figure~\ref{fig:weight_norm_main} depicts the evolution of the actor weight norm throughout training. The baseline maintains a relatively high weight norm across the O2O transition, whereas REFIT produces a sharp reduction at the start of online fine-tuning, consistent with transferring the offline policy behavior into a freshly initialized student. These results show that REFIT changes the parameter scale at the transition and, together with the FAU and SRank diagnostics, provide complementary evidence for its plasticity-restoring effect.

\paragraph{SRank}
\begin{figure}[htbp]
    \vspace{0ex}
    \centering
    \includegraphics[width=\textwidth]{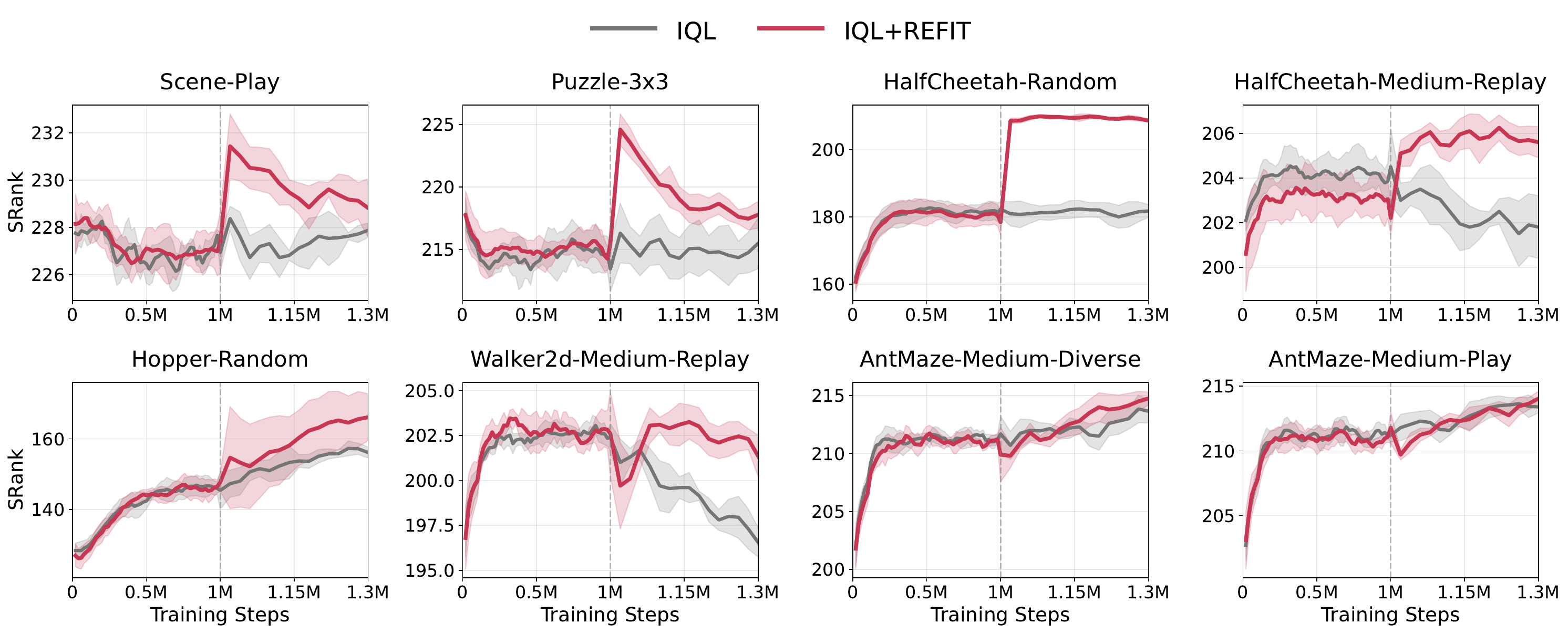}
    \caption{SRank of REFIT and baseline throughout the training process.}
    \label{fig:srank_main}
\end{figure}
\vspace{1ex}

Finally, we evaluate the representational richness using SRank, where a higher value signifies a more diverse and robust feature space~\citep{gulcehre2022empirical, lyle2023understanding}. As shown in Figure~\ref{fig:srank_main}, transferring directly from O2O learning often causes the baseline representations to suffer from representational stagnation, characterized by a stabilized yet severely bounded SRank that fails to expand when encountering new online environments. REFIT increases SRank sharply on several tasks and is comparable on others; the relationship is task-dependent. This preservation of higher-rank representations suggests that the network retains sufficient plasticity and adaptive capacity to capture the novel dynamics encountered during online exploration.

\clearpage

\section{Limitations and Discussion}
\label{appendix:limitations}

While REFIT consistently improves aggregate performance and outperforms the baselines and other plug-in methods on the majority of environments, its primary contribution lies in addressing O2O adaptation from the perspective of network plasticity. 
This perspective is largely orthogonal to existing methods that target other bottlenecks in O2O RL, such as inaccurate value estimation~\citep{shin2025online}. 
A more systematic understanding of how these factors interact, as well as how complementary mechanisms can be jointly incorporated into O2O RL, remains an interesting direction for future work.

\end{document}